%% file: main.tex
\documentclass[10pt,twocolumn,letterpaper]{article}

\usepackage[pagenumbers]{cvpr}

\input{preamble}

\definecolor{cvprblue}{rgb}{0.21,0.49,0.74}
\usepackage[pagebackref,breaklinks,colorlinks,allcolors=cvprblue]{hyperref}
\hypersetup{pdftitle={Resolution as Attention for Efficient Agent Memory Perception}, pdfauthor={Siqi Xiang, Zhipeng Xu, Shengrui Liu, Zulong Chen, Chunyan Miao, Shijian Lu}}

\title{Resolution as Attention for Efficient Agent Memory Perception}

\author{
Siqi Xiang$^{1}$\quad
Zhipeng Xu$^{2}$\quad
Shengrui Liu$^{1}$\quad
Zulong Chen$^{2}$\quad
Chunyan Miao$^{1}$\quad
Shijian Lu$^{1}$\thanks{Corresponding author.}\\[0.4em]
$^{1}$Nanyang Technological University\qquad
$^{2}$Alibaba Group\\[0.4em]
{\raisebox{-0.17em}{\includegraphics[height=1.05em]{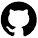}}\hspace{0.35em}\href{https://sxiangag.github.io/resolution-as-attention/}{\texttt{sxiangag.github.io/resolution-as-attention}}}
}

\newcommand{\methodname}{\emph{Resolution as Attention}}
\newcommand{\shortmethodname}{\emph{RAA}}

\begin{document}
\maketitle
\input{section/abstract}
\input{section/introduction}
\input{section/related_work}
\input{section/preliminaries}
\input{section/methodology}
\input{section/experiment}
\input{section/conclusion}
\input{section/acknowledgement}
{
    \small
    \bibliographystyle{ieeenat_fullname}
    \bibliography{main}
}

\input{section/suppl}

\end{document}

%% file: preamble.tex
\usepackage{multirow}
\usepackage{colortbl}

\usepackage{pifont}

\definecolor{tableOursBg}{HTML}{E8F2FC}
\definecolor{tableAvgBg}{HTML}{FFF5DF}
\definecolor{tableTokenBg}{HTML}{F2F2F2}
\definecolor{tableReduction}{HTML}{247A42}
\definecolor{tableSectionBg}{HTML}{EAF1E9}

\definecolor{tableCheck}{HTML}{247A42}
\definecolor{tableCross}{HTML}{B5403A}
\newcommand{\cmark}{\textcolor{tableCheck}{\ding{51}}}
\newcommand{\xmark}{\textcolor{tableCross}{\ding{55}}}

\usepackage[most]{tcolorbox}
\usepackage{listings}
\definecolor{promptFrame}{HTML}{3A6EA5}
\definecolor{promptBack}{HTML}{F7F9FC}
\definecolor{promptTitle}{HTML}{E8F2FC}
\lstdefinestyle{prompt}{
  basicstyle=\ttfamily\fontsize{8}{9}\selectfont, aboveskip=0pt, belowskip=0pt,
  breaklines=true, breakatwhitespace=true,
  columns=fullflexible, keepspaces=true,
  showstringspaces=false,
  literate={–}{{-}}1 {—}{{-}}1 {’}{{'}}1
}
\newtcblisting{promptbox}[1]{
  enhanced, listing only,
  listing options={style=prompt},
  colback=promptBack, colframe=promptFrame,
  coltitle=black, colbacktitle=promptTitle,
  fonttitle=\small\bfseries, title={#1},
  boxrule=0.5pt, arc=2pt,
  left=4pt, right=4pt, top=3pt, bottom=3pt,
  toptitle=2pt, bottomtitle=2pt,
}

%% file: section/abstract.tex
\begin{abstract}

LLM-based agents accumulate extensive interaction histories when solving complex, long-horizon tasks, making efficient memory management critical to their performance and scalability.
While existing memory management methods primarily focus on what to retain or retrieve, they often overlook how much perceptual detail should be allocated to individual past interactions.
We posit that effective agent memory depends not only on the information preserved, but also on the granularity at which each past interaction is perceived.
Motivated by this insight, we propose \textbf{\textit{Resolution as Attention (RAA)}}, a memory perception mechanism that uses visual resolution as a means of attention allocation across interaction histories.
RAA renders past interactions as images and dynamically adjusts their resolution at every step according to their relevance to the agent's upcoming decision, preserving fine-grained information for critical interactions while reducing visual-token consumption for less relevant ones.
To adaptively determine the perceptual fidelity of each interaction, we train a lightweight relevance estimator with teacher-model guidance and further optimize it to preserve the agent's decision consistency at a reduced visual-token cost.
Experiments across multiple tasks and backbones show that RAA reduces token consumption by over 50\% without compromising task performance.
Moreover, RAA can be integrated as a plug-in perception layer with existing memory management methods, yielding further token savings and revealing adaptive perception as a complementary dimension of agent memory.
\end{abstract}

%% file: section/introduction.tex
\section{Introduction}
\label{sec:introduction}

LLM-based agents have demonstrated remarkable capabilities in long-horizon tasks that unfold over multiple rounds of interactions~\cite{react2023,toolformer2023,search-r1-2025,survey-autonomous-agent2024,voyager2024}.
Reaching beyond single-pass answer generation, they autonomously complete complex tasks that once required sustained human effort, such as software engineering~\cite{sweagent2024,openhands2025}, web research~\cite{deepresearcher2025,websailor2025}, and computer use~\cite{osworld2024,ui-tars2025}.
Crucial to this success is \emph{agent memory}, which condenses the ever-growing interaction history into a compact context and thereby enables agents to carry their experience across long horizons coherently~\cite{survey-memory-mechanism-of-large-language-model-based-agents-2025,survey-memory-in-the-age-of-ai-agents2025}.

\begin{figure*}[t]
\centering
\includegraphics[width=0.96\textwidth]{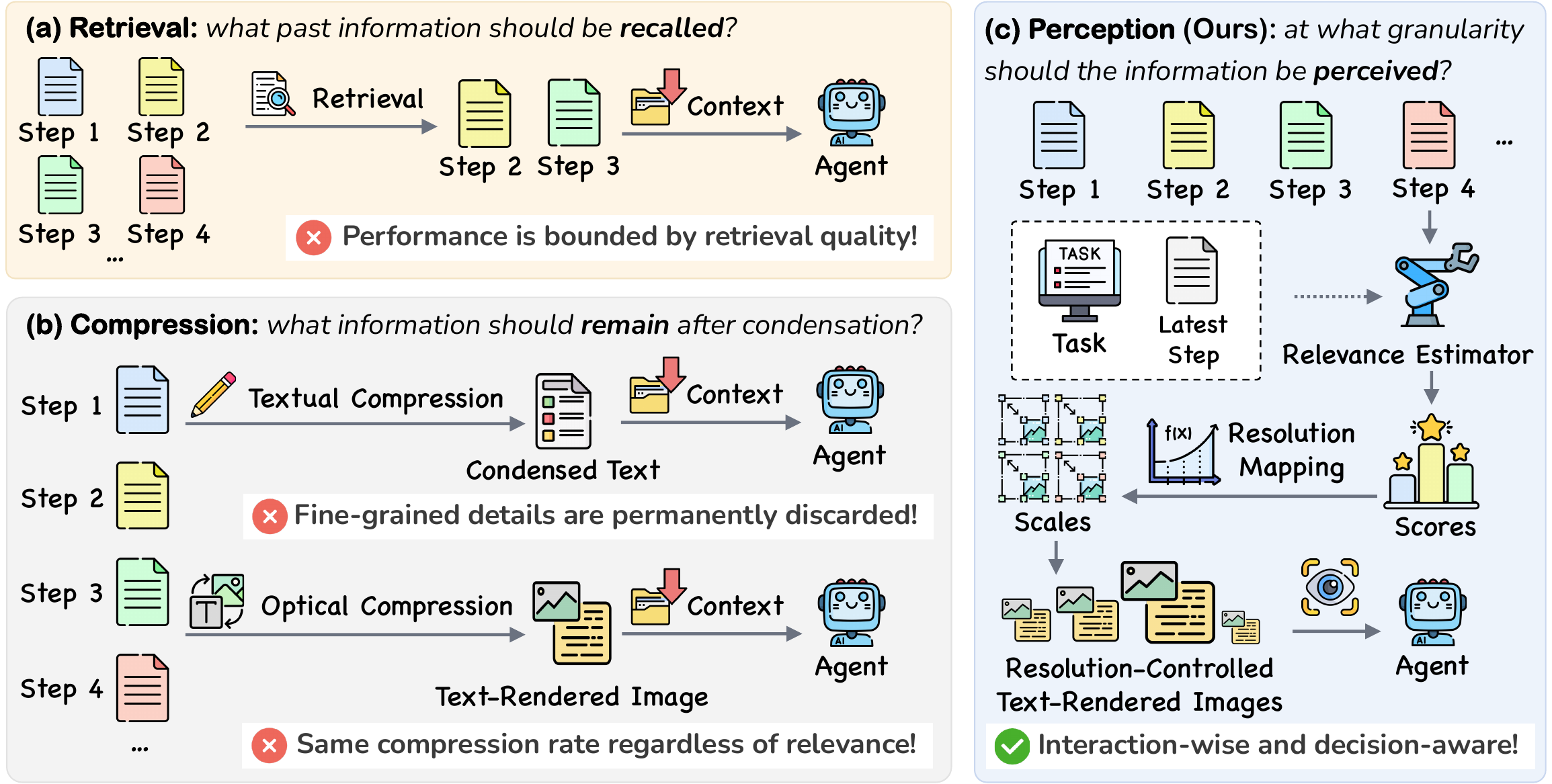}
\caption{
Comparison of \textbf{existing memory management approaches} and \textbf{\methodname{}}.
(a) \emph{Retrieval} is bounded by retrieval quality.
(b) \emph{Textual compression} permanently discards fine-grained details, while \emph{optical compression} applies the same compression rate regardless of relevance.
(c) \shortmethodname{} perceives each interaction at a resolution scaled by its relevance to the upcoming decision.
}
\label{fig:introduction}
\end{figure*}

Given that protracted interactions easily inflate the LLM's context, condensing the history into a concise representation has become a critical challenge for agent memory, with \emph{retrieval} and \emph{compression} emerging as the two predominant approaches.
As shown in \cref{fig:introduction}(a), \emph{retrieval} stores the history in external memory and recalls relevant episodes when needed~\cite{memgpt2023,mem0-2025,amem2025,memorybank2024}.
Despite their effectiveness, these methods are fundamentally bounded by the quality of the retrieved context, as any episode left unrecalled is withheld from the agent~\cite{crag2024,robustretrieval2024}.
As shown in \cref{fig:introduction}(b), \emph{compression}, on the other hand, condenses the accumulated history into a compact form, often through summarization~\cite{resum2025,acon2025} or pruning~\cite{complexitytrap2025,llmlingua-2-2024}.
While keeping the context concise, compression irreversibly abstracts away fine-grained information, which can be critical to subsequent decisions~\cite{complexitytrap2025}.
Recent advances in optical context compression further explore replacing textual content with images, which can be read through fewer visual tokens, as shown in \cref{fig:introduction}(b)~\cite{deepseekocr2025,deepseekocr2-2026,glyph2026,agentocr2026}.
While achieving substantial compression gains, these methods typically compress the history at a uniform rate or according to a fixed policy~\cite{agentocr2026,memocr2026,ocrmemory2026,vera2026}.
Such uniformity leads to a waste of visual tokens on irrelevant interactions, or insufficient detail for critical ones, which hinders the agent's performance.

We posit that the effectiveness of agent memory depends on the granularity at which each past interaction is perceived by the agent, and that this granularity should vary across interactions and evolve as the task progresses.
Building on this insight, we propose \textbf{\methodname{}\emph{~(RAA)}}, a memory perception mechanism that imposes attention on each past interaction by controlling the resolution at which it is read.
Inspired by human vision, in which attention is allocated through the precision of visual perception~\cite{attention-visual-resolution1998}, \methodname{} renders each past interaction as an image and scales its resolution according to its relevance to the upcoming decision.
By dynamically calibrating the resolution of each past interaction, \emph{RAA} efficiently reduces token consumption while preserving the agent's performance.
To make \emph{RAA} compatible with off-the-shelf agents, we decouple resolution allocation from the agent itself.
The resolution is decided by a lightweight module before the agent perceives the memory, and is realized purely by resizing images rendered from the original textual content.
The agent therefore requires no additional training to control the perception process, allowing \emph{RAA} to be plugged into vision-capable agents as a perception layer.

We instantiate \methodname{} with a lightweight relevance estimator fine-tuned from Qwen3-Reranker-0.6B~\cite{qwen3embedding2025}, which judges the relevance of each past interaction to the upcoming decision given the task and the latest interaction.
A learnable normalized sigmoid mapping then translates this relevance into the resolution at which each interaction is perceived.
We train \emph{RAA} in two stages.
In the first stage, the relevance estimator is fine-tuned to match the relevance scores assigned by a teacher model on high-scoring trajectories.
In the second stage, the estimator and the mapping are jointly calibrated with reinforcement learning, which rewards preserving the agent's original decisions with fewer visual tokens.
The agent itself remains frozen during both stages, allowing \emph{RAA} to be trained once and transferred to other agents without further fine-tuning.
Experiments on ClawGym-Bench~\cite{clawgym2026} show that \emph{RAA} reduces history token consumption by up to 57.5\% across all three agents~\cite{qwen3.8-2026,qwen3.6-27b-2026,kimi-k3-2026}, two of which are unseen during training.
It consistently maintains comparable task performance and achieves the lowest tokens per task among all baselines.
We also demonstrate that \emph{RAA}, serving as a memory perception layer, generalizes across existing memory management methods~\cite{acon2025,amem2025,complexitytrap2025}, further reducing their token consumption by up to 54.1\% while preserving task performance.
Our main contributions are summarized as follows:
\begin{itemize}
\item We introduce memory perception as a dimension of agent memory, complementary to retrieval and compression, dictating the granularity at which each past interaction is perceived. We realize efficient memory perception through \methodname{}, which casts the agent's attention over its memory as image resolution. To the best of our knowledge, our approach is the first to allocate perceptual attention across past interactions according to their evolving relevance to the upcoming decision.
\item We design \methodname{} as a lightweight perception layer, comprising a relevance estimator and a normalized sigmoid mapping. Orthogonal to both the agent and its underlying memory, \shortmethodname{} transfers to unseen off-the-shelf agents and composes with existing memory management methods without additional fine-tuning.
\item Extensive experiments on ClawGym-Bench with Qwen-3.8-27B~\cite{qwen3.8-2026}, Qwen-3.6-27B~\cite{qwen3.6-27b-2026}, and Kimi-K3~\cite{kimi-k3-2026} show that \methodname{} substantially reduces token consumption while preserving task performance, outperforming all baselines. Ablations and analyses further validate the effectiveness of each component and training stage, as well as the generalization of \methodname{} to existing memory management methods.
\end{itemize}

%% file: section/related_work.tex
\begin{figure*}[t]
\centering
\includegraphics[width=0.98\textwidth]{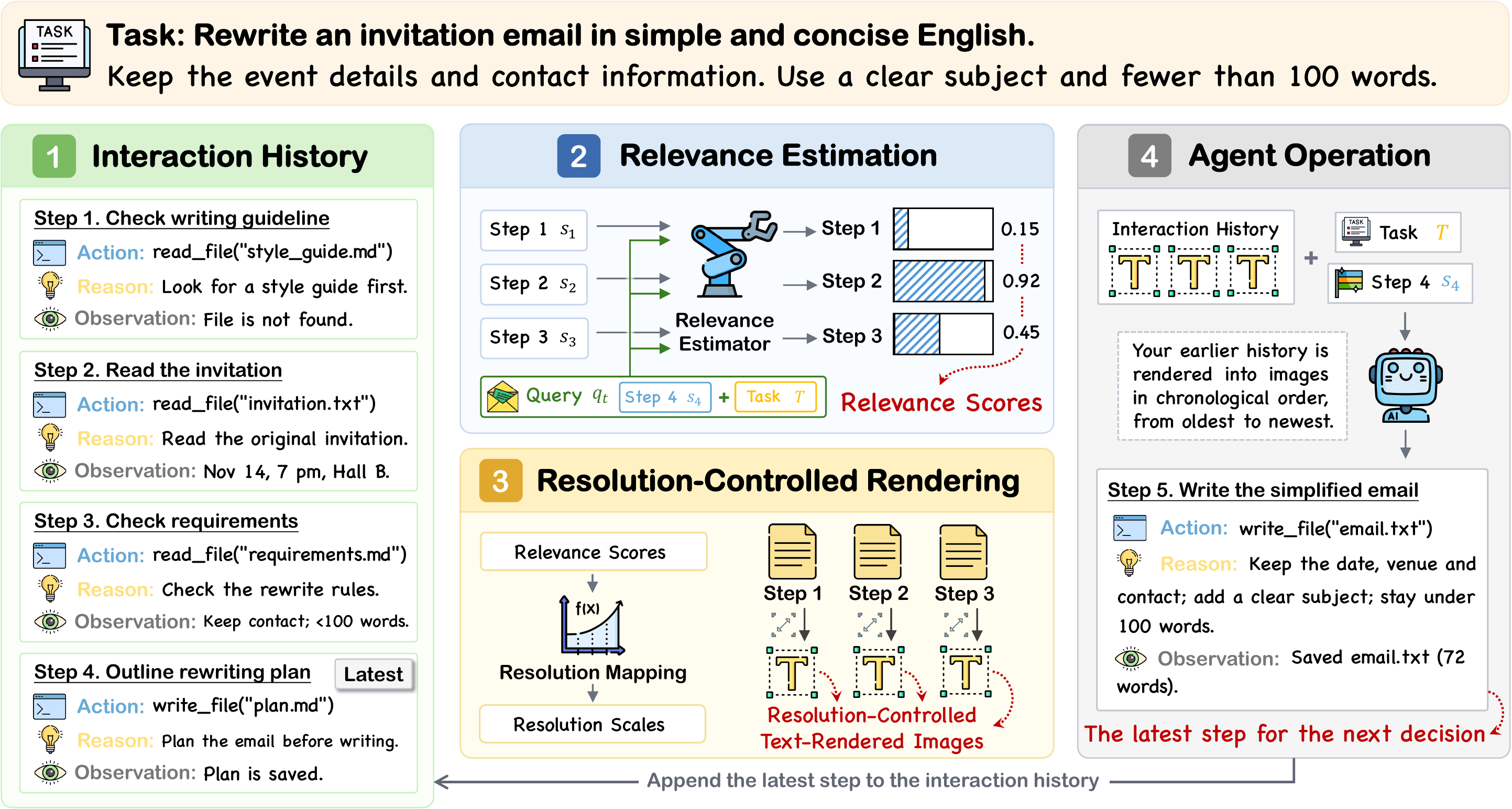}
\caption{
Overview of \methodname{}.
\textbf{Relevance estimation} scores each past interaction against the task and the latest interaction, and \textbf{resolution-controlled rendering} maps these scores to the scales at which the interactions are perceived by the agent.
}
\label{fig:overview}
\end{figure*}

\section{Related Work}
\label{sec:related_work}

\paragraph{Long-horizon LLM agents.}
Leveraging the reasoning and tool-use capabilities of LLMs, LLM agents have advanced to tackling complex tasks through iterative interaction with external environments~\cite{react2023,toolformer2023,survey-autonomous-agent2024}.
Recent efforts have explored the use of agents in software engineering~\cite{sweagent2024,openhands2025}, web research~\cite{deepresearcher2025,websailor2025}, and computer use~\cite{osworld2024,ui-tars2025}, where tasks typically unfold over extended sequences of interactions.
These interactions generate an expanding history of reasoning, actions, and observations that informs subsequent decisions.
The input context thus grows with the interaction horizon, making memory management a central challenge for long-horizon LLM agents~\cite{survey-memory-mechanism-of-large-language-model-based-agents-2025,survey-memory-in-the-age-of-ai-agents2025}.

\paragraph{Agentic memory management.}
Given the finite context window of LLMs, various approaches have been proposed to manage agent memory~\cite{survey-memory-mechanism-of-large-language-model-based-agents-2025,survey-memory-in-the-age-of-ai-agents2025}.
These methods primarily follow two paradigms, namely \textit{retrieval} and \textit{compression}.
Retrieval-based methods store past interactions in an external memory and recall relevant episodes into the context on demand.
MemGPT~\cite{memgpt2023} organizes memory into an operating-system-like hierarchy, while Mem0~\cite{mem0-2025} and A-MEM~\cite{amem2025} store memory as interconnected nodes.
Their effectiveness, however, hinges on the retrieval quality, as episodes that are unrecalled remain invisible to the agent~\cite{crag2024,robustretrieval2024}.
Compression-based methods instead condense the interaction history into a compact representation.
ACON~\cite{acon2025} and ReSum~\cite{resum2025} periodically summarize the history, while observation masking~\cite{complexitytrap2025} and LLMLingua-style~\cite{llmlingua2023,llmlingua-2-2024,longllmlingua2024} methods prune earlier history.
However, these methods irreversibly discard fine-grained information that may be critical to subsequent decisions~\cite{complexitytrap2025}.
\textit{Retrieval} and \textit{compression} determine which content is recalled or retained, whereas \methodname{} adjusts the granularity at which the memory is perceived.

\paragraph{Optical context compression.}
In recent research, optical context compression has emerged as a promising direction for long-context compression.
Utilizing the Optical Character Recognition (OCR) capabilities of vision-language models, it renders text into images that can be read through fewer visual tokens.
DeepSeek-OCR~\cite{deepseekocr2025,deepseekocr2-2026} and Glyph~\cite{glyph2026} show that text can be decoded from rendered images at substantial compression ratios.
AgentOCR~\cite{agentocr2026} extends this to agent memory, training the agent to control the compression rate and read from the rendered image.
MemOCR~\cite{memocr2026} trains a model to rewrite its memory into rich text that emphasizes crucial information through layout.
OCR-Memory~\cite{ocrmemory2026} and VERA~\cite{vera2026} further explore downsampling older interactions by recency.
However, these methods either compress the memory at a uniform rate or by a fixed recency policy.
\methodname{} instead treats memory reading as perception, using image resolution as a means to control the granularity at which each past interaction is perceived.
This perceptual perspective opens a dimension of agent memory orthogonal to existing retrieval and compression methods by making memory reading interaction-wise, decision-aware, and non-destructive.

%% file: section/preliminaries.tex
\section{Preliminaries}
\label{sec:preliminaries}

We first formalize the long-horizon agent setting in which the interaction history accumulates over time (\cref{sec:problem_formulation}), then introduce text-rendered memory, where the resolution of each image determines its perceptual fidelity (\cref{sec:memory_and_resolution}).

\subsection{Problem Formulation}
\label{sec:problem_formulation}

Consider an agent solving long-horizon tasks through multiple rounds of interactions.
Given the task description $D$, at each step $t$, the agent $\pi$ generates reasoning content $r_t$ followed by an action $a_t$, and the environment responds with an observation $o_t$.
We denote the $i$-th interaction as $h_i = (r_i, a_i, o_i)$ and the history before step $t$ as $\mathcal{H}_t = (h_1, \dots, h_{t-1})$.
The agent produces the next reasoning content and action conditioned on the task and the history:
\begin{equation}
    (r_t, a_t) \sim \pi(\cdot \mid D, \mathcal{H}_t).
    \label{eq:policy}
\end{equation}
Since every interaction is appended to $\mathcal{H}_t$, the context inflates as the task proceeds.
In this work, we treat $\pi$ as a frozen vision-capable agent and study how to present $\mathcal{H}_t$ to it at a lower token cost according to the relevance of each interaction $h_i$ to the upcoming decision at step $t$.

\subsection{Text-Rendered Memory}
\label{sec:memory_and_resolution}

Rendering text into images makes the detail of each interaction adjustable through resolution.
We render each past interaction $h_i$ into a canonical image $I_i = \mathcal{R}(h_i)$ of width $W_i$ and height $L_i$, which remains unchanged throughout the task.
When reading the memory, the agent perceives $I_i$ at a resolution scale $s_i \in (0, 1]$, which resizes both sides of the canonical image by a factor of $s_i$:
\begin{equation}
    \tilde{I}_i = \mathrm{Resize}(I_i, s_i).
    \label{eq:resize}
\end{equation}
Approximately, for a vision encoder that maps each $P \times P$ patch to a single visual token, the resized image consumes:
\begin{equation}
    N_i(s_i) = \left\lceil \frac{s_i W_i}{P} \right\rceil \left\lceil \frac{s_i L_i}{P} \right\rceil \approx s_i^2 \, N_i(1)
    \label{eq:tokens}
\end{equation}
visual tokens.
The number of visual tokens for an interaction therefore decreases quadratically with its scale, while its detail becomes harder to perceive at lower scales.
The memory presented to the agent is the set of resized images:
\begin{equation}
    \mathcal{M}_t = \big\{ \mathrm{Resize}(I_i, s_i^{(t)}) \big\}_{i=1}^{t-1},
    \label{eq:memory}
\end{equation}
where $s_i^{(t)}$ is the scale of $h_i$ at step $t$.
Each interaction can thus be perceived anew at every step as the task proceeds.

%% file: section/methodology.tex
\section{Resolution as Attention}
\label{sec:methodology}

We now introduce \methodname{}, a memory perception mechanism that imposes attention on each past interaction by controlling the resolution at which it is read.
We first present the overall framework (\cref{sec:overview}).
As shown in \cref{fig:overview}, \methodname{} estimates the relevance of each past interaction to the upcoming decision (\cref{sec:relevance}), maps the relevance to a resolution scale, and resizes the canonical images of the past interactions at these scales to form the memory that the agent perceives (\cref{sec:rendering}).
Finally, we describe our two-stage training procedure that learns the relevance estimator and the resolution mapping while keeping the agent intact (\cref{sec:learning}).

\subsection{Framework Overview}
\label{sec:overview}

As shown in \cref{fig:overview}, \methodname{} dynamically adjusts the perceptual granularity of the interaction history $\mathcal{H}_t$ before every decision.
Specifically, prior to generating $r_t$ and $a_t$, a relevance estimator $f_\theta$ forms a query by combining the task description $D$ and the latest interaction $h_{t-1}$.
It then matches this query against each past interaction $h_i$ to estimate its relevance to the upcoming decision:
\begin{equation}
p_i^{(t)} = f_\theta\big(D, h_{t-1}, h_i\big) \in (0, 1).
\label{eq:overview_relevance}
\end{equation}
To translate the relevance $p_i^{(t)}$ into the granularity at which each interaction is perceived, a monotone mapping $g_\phi$ is applied to convert each relevance score into a resolution scale:
\begin{equation}
s_i^{(t)} = g_\phi\big(p_i^{(t)}\big) \in (0, 1],
\label{eq:overview_mapping}
\end{equation}
at which the canonical images $I_i$ of the corresponding interactions are resized to form the memory $\mathcal{M}_t$ shown in \cref{eq:memory}.
The agent then perceives $\mathcal{M}_t$ and generates the next reasoning content $r_t$ and action $a_t$ accordingly:
\begin{equation}
(r_t, a_t) \sim \pi\big(\cdot \mid D, \mathcal{M}_t\big).
\label{eq:overview_policy}
\end{equation}
The environment returns an observation $o_t$, and the resulting interaction $h_t$ is appended to the interaction history.
The resolution of each interaction is thus adapted based on its relevance to the upcoming decision as the task unfolds.

\methodname{} does not impose a global token budget constraint on perception, unlike the softmax in attention, which distributes a fixed total weight across keys.
Before the memory is perceived, the amount of information required by the upcoming decision remains unknown, so a preset budget may either truncate critical details or retain redundant ones.
We therefore scale each interaction independently according to its relevance score, allowing the total visual tokens to vary with the demand of each decision.

\subsection{Relevance Estimation}
\label{sec:relevance}

The relevance estimator $f_\theta$ anticipates the relevance of each past interaction to the upcoming decision.
Concretely, we instantiate $f_\theta$ as a lightweight reranker, which takes the task description $D$ and the latest interaction $h_{t-1}$ as the query and scores each past interaction against the query.
The relevance is given by the probability of an affirmative judgment:
\begin{equation}
p_i^{(t)} = \sigma\big(z_{\text{yes}} - z_{\text{no}}\big),
\label{eq:relevance}
\end{equation}
where $z_{\text{yes}}$ and $z_{\text{no}}$ are the logits of the tokens ``yes'' and ``no'', respectively, and $\sigma$ denotes the sigmoid function.

Much as a token being generated in an autoregressive model is conditioned on preceding tokens, we regard the upcoming decision as conditioned on the past interactions.
Following this analogy, the task description $D$ and the latest interaction $h_{t-1}$ jointly constitute the query for the relevance estimator, while each past interaction $h_i$ serves as a key and its canonical image $I_i$ as the corresponding value.
The relevance score $p_i^{(t)}$ is therefore a measure of the dependence of the upcoming decision on the information in $h_i$.
Each interaction is scored in isolation, so that its relevance is not diluted by the presence of other interactions in long-horizon tasks.
As the query is renewed with the latest interaction at every step, the relevance of each past interaction evolves with the progress of the task.

\subsection{Resolution-Controlled Rendering}
\label{sec:rendering}

Resolution-controlled rendering is responsible for turning relevance into the images that the agent perceives.
It first converts each relevance score into a resolution scale through a learnable monotone mapping $g_\phi$, parameterized as a normalized sigmoid function:
\begin{equation}
g_\phi(p) = \frac{\sigma\big(k(p - b)\big) - \sigma(-kb)}{\sigma\big(k(1 - b)\big) - \sigma(-kb)},
\label{eq:mapping}
\end{equation}
where $\phi = \{k, b\}$ are learnable parameters, with $k > 0$ controlling the sharpness of the transition and $b \in (0, 1)$ determining the midpoint of the normalized curve.
This normalization ensures $g_\phi(0) = 0$ and $g_\phi(1) = 1$, so that the mapping spans the full range of scales.
Being monotone, it preserves the ranking induced by relevance scores while reconciling the relevance distribution with that of the resolution scales the agent requires to perceive each interaction.

Given the scale $s_i^{(t)}$, \methodname{} resizes the canonical image $I_i$ following \cref{eq:resize}.
The resized images of all past interactions together constitute the memory $\mathcal{M}_t$ in \cref{eq:memory}.
Each resizing starts from the canonical image rather than its last rendering, so that every interaction is perceived at a resolution commensurate with its current relevance.
The canonical images are thus preserved throughout the task, while their perception adapts at every step.

\subsection{Learning Relevance Estimation and Resolution Mapping}
\label{sec:learning}

We learn the relevance estimator $f_\theta$ and the mapping $g_\phi$ in two stages, namely relevance supervision (\textbf{Stage I}) and resolution reinforcement (\textbf{Stage II}).
Relevance supervision distills the relevance judgments from a teacher model into the estimator.
Resolution reinforcement then calibrates the relevance estimator and the resolution mapping to preserve the agent's original decisions with fewer visual tokens.

\paragraph{Relevance supervision (Stage I).}
A pretrained reranker such as Qwen3-Reranker-0.6B~\cite{qwen3embedding2025} typically judges whether a passage matches a query in semantic meaning, whereas \methodname{} requires the extent to which the upcoming decision depends on each past interaction.
These two notions of relevance are related yet distinct, as a past interaction may be semantically related to the query but exert little influence on the decision, and vice versa.
We therefore distill such decision relevance from a teacher model into the lightweight estimator $f_\theta$.

To obtain supervision signals that reflect decision relevance, we collect decision histories from high-scoring agent trajectories on long-horizon tasks.
At every step of these trajectories, the teacher model is given the task description $D$ and the history $\mathcal{H}_t$, anticipates the next action, and rates the dependence $\hat{p}_i^{(t)} \in [0, 1]$ of the anticipated action on each past interaction.
The relevance estimator is then fine-tuned to regress these ratings with a density-weighted mean squared error loss:
\begin{equation}
\mathcal{L}_{\text{sup}}(\theta) = \mathbb{E}_{t, i}\Big[w_i^{(t)}\big(p_i^{(t)} - \hat{p}_i^{(t)}\big)^2\Big],
\label{eq:sup}
\end{equation}
where $p_i^{(t)}$ is the relevance score predicted by $f_\theta$ and $\hat{p}_i^{(t)}$ is the rating given by the teacher model, with $w_i^{(t)}$ reweighting each pair inversely to the density of its rating.

Relevance supervision grounds the estimator in decision relevance before resolution is introduced.
Such grounding is crucial, as the effectiveness of resolution reinforcement builds upon the ability of the relevance estimator to single out interactions on which the upcoming decision depends.

\paragraph{Resolution reinforcement (Stage II).}
Decision relevance indicates the importance of each past interaction, yet the resolution each interaction requires is determined by the perceptual capability of the agent.
We therefore calibrate the estimator and the resolution mapping jointly with reinforcement learning, guided by feedback from the frozen agent on how faithfully it preserves its original decisions.

Specifically, we treat the composition $g_\phi \circ f_\theta$ as a stochastic policy over resolution scales.
For a past interaction $h_i$ at step $t$, its predicted scale $s_i^{(t)}$ serves as the center of a logit-normal distribution from which we sample a scale $s_{i,j}$ to resize the canonical image $I_i$:
\begin{equation}
s_{i,j} = \sigma\big(\operatorname{logit} s_i^{(t)} + \tau \epsilon_j\big), \quad \epsilon_j \sim \mathcal{N}(0, 1),
\label{eq:sampling}
\end{equation}
where $\tau$ is the temperature parameter that controls the spread of the distribution.
Such sampling is repeated $K$ times while all other interactions keep their predicted scales, yielding $K$ memories $\{\mathcal{M}_t^{(j)}\}_{j=1}^{K}$ that differ in the resolution of $h_i$.
We then feed each memory to the frozen agent and reward the corresponding scale for preserving the agent's original action $a_t$ while saving visual tokens:
\begin{equation}
\thickmuskip=3mu
R_j = \underbrace{\bar{\ell}\big(a_t \mid D, \mathcal{M}_t^{(j)}\big) - \bar{\ell}\big(a_t \mid D, \mathcal{H}_t\big)}_{\text{decision fidelity}} + \underbrace{\lambda\big(1 - s_{i,j}^2\big)}_{\text{token saving}},
\label{eq:reward}
\end{equation}
where $\bar{\ell}$ denotes the mean token log-likelihood of the action under the frozen agent, $\mathcal{H}_t$ denotes the full textual history that serves as the reference, and $1 - s_{i,j}^2$ approximates the fraction of visual tokens saved on $h_i$ by \cref{eq:tokens}.
The cost weight $\lambda$ governs the trade-off between decision fidelity and token saving, with larger values favoring fewer tokens.

We optimize the estimator and the mapping using group relative policy optimization~\cite{deepseekmath2024} over the $K$ candidates:
\begin{equation}
\mathcal{J}(\theta, \phi) = \mathbb{E}\Big[\tfrac{1}{K}\textstyle\sum_{j=1}^{K} A_j \log q_{\theta, \phi}\big(s_{i,j}\big)\Big],
\label{eq:grpo}
\end{equation}
where $A_j$ is the advantage obtained by normalizing $R_j$ within the $K$ candidates, and $q_{\theta, \phi}$ is the logit-normal distribution in \cref{eq:sampling}.
Since the candidates differ only in the scale of $h_i$, each advantage attributes the change in decision fidelity and token saving precisely to that interaction.

%% file: section/experiment.tex
\section{Experiments}
\label{sec:experiments}

In this section, we conduct extensive experiments to evaluate \methodname{}.
First, we describe the experimental setup (\cref{sec:experiments:setup}) and present the main results (\cref{sec:experiments:main}).
Second, we ablate its components and training stages (\cref{sec:experiments:relevance,sec:experiments:training}).
Third, we analyze the cost weight $\lambda$ (\cref{sec:experiments:lambda}) and its complementarity with existing memory management methods (\cref{sec:experiments:complementarity}).

\subsection{Experimental Setup}
\label{sec:experiments:setup}

\paragraph{Benchmarks and metrics.}
We primarily evaluate \methodname{} on ClawGym-Bench~\cite{clawgym2026}, a long-horizon benchmark comprising 200 agentic tasks across six categories.
Following its official evaluation protocol, we score each task by combining code-based verification with rubric-based assessment and report the hybrid average score.
To quantify token cost, we report the average values of two metrics across all tasks: (1) the total number of history tokens consumed and (2) the peak number of history tokens in a single agent call.
We further evaluate \methodname{} on AppWorld~\cite{appworld2024} in \cref{sec:supp:appworld}.

\paragraph{Baselines.}
We compare \methodname{} with representative baselines that expose the memory to the agent through textual or visual perception.
Textual baselines retain varying portions of the history.
Specifically, Vanilla exposes the full history to the agent, FIFO~\cite{sweagent2024} restricts access to the five most recent rounds of the history, and LongLLMLingua~\cite{longllmlingua2024} provides a pruned textual context conditioned on the task description and the latest round.
Visual baselines instead present the history as rendered images.
OCR-Memory~\cite{ocrmemory2026} preserves the resolution of the five most recent rounds while halving that of earlier rounds, whereas AgentOCR~\cite{agentocr2026} allows the agent to select a global compression factor for the rendered history.
All methods run within the OpenClaw framework~\cite{openclaw2026}, with further implementation details provided in \cref{sec:supp:baselines}.

\paragraph{Implementation details.}
We train \methodname{} with Qwen-3.8-27B~\cite{qwen3.8-2026} as the frozen agent and initialize the relevance estimator from Qwen3-Reranker-0.6B~\cite{qwen3embedding2025}.
For relevance supervision, we use 12,000 decision histories collected from 1,853 ClawGym training tasks, with interaction-level relevance ratings generated by Qwen-3.8-27B.
Resolution reinforcement then jointly optimizes the estimator and the resolution mapping for 150 updates.
At inference time, each past interaction is resized to its predicted scale, except for the latest one, which forms the query and is kept as text.
To assess cross-agent transfer, we additionally deploy the trained module on Qwen-3.6-27B~\cite{qwen3.6-27b-2026} and Kimi-K3~\cite{kimi-k3-2026} without further fine-tuning.
All evaluation backbones remain frozen and use a temperature of 0.
Further details are provided in \cref{sec:supp:implementation}.

\begin{table*}[t]
\centering
\caption{
Results on \textbf{ClawGym-Bench}.
Higher scores and lower token costs are better.
Bold and underlined values indicate the best and second-best results
excluding Vanilla within each backbone, respectively.
Percentages indicate reductions relative to Vanilla.
}
\label{tab:clawgym-main}
\setlength{\tabcolsep}{4pt}
\renewcommand{\arraystretch}{1.12}
\resizebox{\textwidth}{!}{%
\begin{tabular}{
@{}ll*{6}{c}
!{\hspace{3pt}}
>{\columncolor{tableAvgBg}}c
!{\hspace{3pt}}
*{2}{>{\columncolor{tableTokenBg}}c}
@{}
}
\toprule
\textbf{Backbone}
& \textbf{Method}
& \begin{tabular}[c]{@{}c@{}}Product.\\\& Collab.\end{tabular}
& \begin{tabular}[c]{@{}c@{}}Systems\\\& Auto.\end{tabular}
& \begin{tabular}[c]{@{}c@{}}Analysis\\\& Reason.\end{tabular}
& \begin{tabular}[c]{@{}c@{}}Content\\\& Domain\end{tabular}
& \begin{tabular}[c]{@{}c@{}}Planning\\\& Knowl.\end{tabular}
& \begin{tabular}[c]{@{}c@{}}Software\\Dev.\end{tabular}
& \textbf{Avg.}
& \textbf{Tokens/Task}
& \textbf{Peak Tokens} \\
\midrule

\multirow{6}{*}{Qwen-3.8-27B}
& Vanilla
& 78.82 & 67.16 & 82.43 & 83.10 & 79.44 & 83.23
& 78.23
& 89733
& 12561 \\
\cmidrule(l){2-11}

& FIFO
& 80.69 & 70.16 & 78.99 & 80.06 & 77.89 & 82.50
& \underline{77.95}
& \underline{48825}\,{\scriptsize\textcolor{tableReduction}{($\downarrow$ 45.6\%)}}
& 8451\,{\scriptsize\textcolor{tableReduction}{($\downarrow$ 32.7\%)}} \\

& LongLLMLingua
& 79.00 & 66.18 & 76.62 & 83.54 & 80.06 & 88.61
& 77.87
& 62035\,{\scriptsize\textcolor{tableReduction}{($\downarrow$ 30.9\%)}}
& 7024\,{\scriptsize\textcolor{tableReduction}{($\downarrow$ 44.1\%)}} \\

& OCR-Memory
& 64.90 & 62.37 & 64.70 & 62.24 & 66.18 & 70.77
& 64.86
& 67921\,{\scriptsize\textcolor{tableReduction}{($\downarrow$ 24.3\%)}}
& \underline{6500}\,{\scriptsize\textcolor{tableReduction}{($\downarrow$ 48.2\%)}} \\

& AgentOCR
& 74.62 & 68.40 & 79.88 & 83.32 & 77.96 & 85.21
& 77.21
& 69693\,{\scriptsize\textcolor{tableReduction}{($\downarrow$ 22.3\%)}}
& 8266\,{\scriptsize\textcolor{tableReduction}{($\downarrow$ 34.2\%)}} \\

\rowcolor{tableOursBg}
\cellcolor{white} & \textbf{RAA (Ours)}
& 77.51 & 70.18 & 79.30 & 84.17 & 81.03 & 85.62
& \cellcolor{tableAvgBg}\textbf{78.69}
& \cellcolor{tableTokenBg}\textbf{38169}\,{\scriptsize\textcolor{tableReduction}{\textbf{($\boldsymbol{\downarrow}$ 57.5\%)}}}
& \cellcolor{tableTokenBg}\textbf{6447}\,{\scriptsize\textcolor{tableReduction}{\textbf{($\boldsymbol{\downarrow}$ 48.7\%)}}} \\
\midrule

\multirow{6}{*}{Qwen-3.6-27B}
& Vanilla
& 68.85 & 64.15 & 67.31 & 71.80 & 70.11 & 67.32
& 67.98
& 64456
& 9284 \\
\cmidrule(l){2-11}

& FIFO
& 66.91 & 60.23 & 65.10 & 68.97 & 68.15 & 63.72
& 65.24
& 40607\,{\scriptsize\textcolor{tableReduction}{($\downarrow$ 37.0\%)}}
& 6674\,{\scriptsize\textcolor{tableReduction}{($\downarrow$ 28.1\%)}} \\

& LongLLMLingua
& 68.46 & 61.78 & 67.82 & 69.08 & 68.32 & 68.98
& \underline{67.08}
& 39415\,{\scriptsize\textcolor{tableReduction}{($\downarrow$ 38.8\%)}}
& 5764\,{\scriptsize\textcolor{tableReduction}{($\downarrow$ 37.9\%)}} \\

& OCR-Memory
& 56.73 & 55.75 & 63.02 & 60.40 & 62.57 & 67.03
& 60.19
& \underline{35719}\,{\scriptsize\textcolor{tableReduction}{($\downarrow$ 44.6\%)}}
& \textbf{3987}\,{\scriptsize\textcolor{tableReduction}{\textbf{($\boldsymbol{\downarrow}$ 57.1\%)}}} \\

& AgentOCR
& 64.61 & 62.57 & 65.56 & 68.18 & 69.90 & 74.04
& 66.71
& 55781\,{\scriptsize\textcolor{tableReduction}{($\downarrow$ 13.5\%)}}
& 6148\,{\scriptsize\textcolor{tableReduction}{($\downarrow$ 33.8\%)}} \\

\rowcolor{tableOursBg}
\cellcolor{white} & \textbf{RAA (Ours)}
& 69.35 & 58.43 & 70.92 & 71.52 & 69.91 & 74.17
& \cellcolor{tableAvgBg}\textbf{68.31}
& \cellcolor{tableTokenBg}\textbf{31230}\,{\scriptsize\textcolor{tableReduction}{\textbf{($\boldsymbol{\downarrow}$ 51.5\%)}}}
& \cellcolor{tableTokenBg}\underline{4512}\,{\scriptsize\textcolor{tableReduction}{($\downarrow$ 51.4\%)}} \\
\midrule

\multirow{6}{*}{Kimi-K3}
& Vanilla
& 79.55 & 71.35 & 80.61 & 84.53 & 82.85 & 85.88
& 79.93
& 73183
& 9419 \\
\cmidrule(l){2-11}

& FIFO
& 79.30 & 68.53 & 80.52 & 86.95 & 81.89 & 83.29
& 79.16
& 43348\,{\scriptsize\textcolor{tableReduction}{($\downarrow$ 40.8\%)}}
& 7222\,{\scriptsize\textcolor{tableReduction}{($\downarrow$ 23.3\%)}} \\

& LongLLMLingua
& 79.40 & 68.13 & 79.90 & 86.20 & 81.26 & 84.52
& 78.95
& 40842\,{\scriptsize\textcolor{tableReduction}{($\downarrow$ 44.2\%)}}
& 6116\,{\scriptsize\textcolor{tableReduction}{($\downarrow$ 35.1\%)}} \\

& OCR-Memory
& 78.80 & 72.35 & 77.94 & 82.79 & 80.07 & 80.00
& 78.17
& \underline{35722}\,{\scriptsize\textcolor{tableReduction}{($\downarrow$ 51.2\%)}}
& \textbf{5051}\,{\scriptsize\textcolor{tableReduction}{\textbf{($\boldsymbol{\downarrow}$ 46.4\%)}}} \\

& AgentOCR
& 78.98 & 68.80 & 80.42 & 86.44 & 84.57 & 86.02
& \underline{79.77}
& 56563\,{\scriptsize\textcolor{tableReduction}{($\downarrow$ 22.7\%)}}
& 8613\,{\scriptsize\textcolor{tableReduction}{($\downarrow$ 8.6\%)}} \\

\rowcolor{tableOursBg}
\cellcolor{white} & \textbf{RAA (Ours)}
& 78.30 & 71.97 & 81.58 & 86.62 & 82.01 & 84.39
& \cellcolor{tableAvgBg}\textbf{79.95}
& \cellcolor{tableTokenBg}\textbf{32258}\,{\scriptsize\textcolor{tableReduction}{\textbf{($\boldsymbol{\downarrow}$ 55.9\%)}}}
& \cellcolor{tableTokenBg}\underline{5400}\,{\scriptsize\textcolor{tableReduction}{($\downarrow$ 42.7\%)}} \\

\bottomrule
\end{tabular}%
}
\end{table*}

\begin{table}[t]
\centering
\caption{
Ablation of \textbf{relevance estimation} and \textbf{resolution-controlled rendering} on ClawGym-Bench using Qwen-3.8-27B.
}
\label{tab:ablation-relevance-rendering}
\setlength{\tabcolsep}{3pt}
\renewcommand{\arraystretch}{1.12}
\resizebox{\columnwidth}{!}{%
\begin{tabular}{
@{}lc
!{\hspace{3pt}}
>{\columncolor{tableAvgBg}}c
!{\hspace{3pt}}
*{2}{>{\columncolor{tableTokenBg}}c}
@{}
}
\toprule
\textbf{Method} & \textbf{Setting}
& \textbf{Avg.}
& \textbf{Tokens/Task}
& \textbf{Peak Tokens} \\
\midrule

\multicolumn{5}{@{}l}{
    \cellcolor{tableSectionBg}
    \strut\textit{Effect of Relevance Estimation}
} \\
\addlinespace[3pt]

\multirow{3}{*}{\shortstack[l]{RAA w/o\\Relevance Estimation}}
& $s=0.70$ & 75.81 & \textbf{36233} & \textbf{5744} \\
& $s=0.85$ & 76.17 & 38676 & 6226 \\
& $s=1.00$ & \underline{77.22} & 47757 & 7031 \\

\rowcolor{tableOursBg}
\textbf{RAA (Ours)}
& --
& \cellcolor{tableAvgBg}\textbf{78.69}
& \cellcolor{tableTokenBg}\underline{38169}
& \cellcolor{tableTokenBg}\underline{6447} \\

\midrule
\multicolumn{5}{@{}l}{
    \cellcolor{tableSectionBg}
    \strut\textit{Effect of Resolution-Controlled Rendering}
} \\
\addlinespace[3pt]

\multirow{3}{*}{\shortstack[l]{RAA w/o\\Rendering}}
& $k=3$ & 77.63 & \underline{42619} & \underline{7929} \\
& $k=4$ & 78.56 & 49767 & 8747 \\
& $k=5$ & \textbf{78.98} & 61149 & 9705 \\

\rowcolor{tableOursBg}
\textbf{RAA (Ours)}
& --
& \cellcolor{tableAvgBg}\underline{78.69}
& \cellcolor{tableTokenBg}\textbf{38169}
& \cellcolor{tableTokenBg}\textbf{6447} \\

\bottomrule
\end{tabular}%
}
\end{table}

\subsection{Main Results}
\label{sec:experiments:main}

We report the comprehensive results of \methodname{} and all baselines on ClawGym-Bench in \cref{tab:clawgym-main}.
Across all backbones, \methodname{} consistently achieves the lowest tokens per task while maintaining task performance comparable to Vanilla, reducing the history tokens per task by 57.5\%, 51.5\%, and 55.9\% on Qwen-3.8-27B, Qwen-3.6-27B, and Kimi-K3, respectively.
In comparison, FIFO, LongLLMLingua, and AgentOCR consume 26.2\% to 82.6\% more tokens per task than \methodname{}, while achieving lower average scores on all backbones.
Since \methodname{} is trained only with Qwen-3.8-27B, the consistent gains on the other two backbones without further fine-tuning demonstrate its transferability and robustness as a perception layer across different agents.
Notably, OCR-Memory reaches slightly lower peak tokens than \methodname{} on Qwen-3.6-27B and Kimi-K3, as its recency-based downscaling compresses earlier interactions regardless of their relevance.
However, this can obscure critical details, causing OCR-Memory to take 7.0\% and 14.0\% more steps and consume 14.4\% and 10.7\% more tokens per task than \methodname{} on these two backbones.
Moreover, with a 0.6B estimator and canonical images rendered only once, relevance estimation and rendering account for only 2.51\% of the end-to-end latency, as detailed in \cref{sec:supp:latency}.

\subsection{Ablation on Relevance Estimation and Resolution-Controlled Rendering}
\label{sec:experiments:relevance}

We ablate the two components of \methodname{} with Qwen-3.8-27B and report the results in \cref{tab:ablation-relevance-rendering}.
For relevance estimation, we remove the estimator and render all past interactions at a uniform scale $s$.
Decreasing $s$ from 1.00 to 0.70 reduces the tokens per task from 47.8K to 36.2K at the cost of a drop in average score from 77.22 to 75.81, as critical details become illegible at a uniformly low resolution.
In contrast, \methodname{} adjusts the resolution of each interaction according to its relevance, achieving a higher average score of 78.69 with 38.2K tokens per task.
It is worth noting that, although rendering at full resolution reduces the tokens per task of Vanilla by 46.8\%, it lowers the average score by 1.01 points, whereas relevance estimation further reduces the tokens by 20.1\% while preserving a score comparable to Vanilla.
For resolution-controlled rendering, we replace rendered images with textual content, keeping the top-$k$ past interactions ranked by the relevance estimator as text and discarding the rest.
Increasing $k$ from 3 to 5 raises the average score from 77.63 to 78.98 but inflates the tokens per task from 42.6K to 61.1K, as every retained interaction is kept in full text.
By comparison, \methodname{} achieves an average score of 78.69 with only 38.2K tokens per task.
The slight performance advantage at $k=5$ stems from recognition errors introduced when reading text from rendered images.

\subsection{Ablation on Two-Stage Training}
\label{sec:experiments:training}

\begin{figure}[t]
\centering
\begin{subfigure}[t]{0.49\linewidth}
\centering
\includegraphics[width=\linewidth]{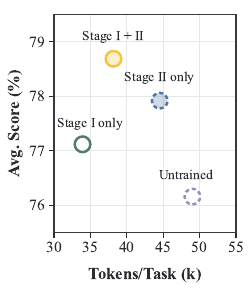}
\caption{Performance and token cost}
\label{fig:ablation-training-a}
\end{subfigure}
\hfill
\begin{subfigure}[t]{0.49\linewidth}
\centering
\includegraphics[width=\linewidth]{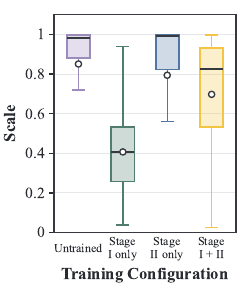}
\caption{Distribution of predicted scales}
\label{fig:ablation-training-b}
\end{subfigure}
\caption{
Ablation on two-stage training with Qwen-3.8-27B.
\textbf{Untrained} uses the pretrained reranker and the initial mapping, \textbf{Stage I} and \textbf{Stage II} apply relevance supervision and resolution reinforcement, respectively, and \textbf{Stage I + II} applies both.}
\label{fig:ablation-training}
\end{figure}

\begin{figure}[t]
\centering
\begin{subfigure}[t]{0.49\linewidth}
\centering
\includegraphics[width=\linewidth]{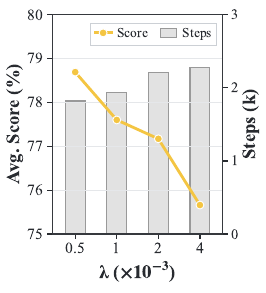}
\caption{Performance and number of steps}
\label{fig:ablation-lambda-a}
\end{subfigure}
\hfill
\begin{subfigure}[t]{0.49\linewidth}
\centering
\includegraphics[width=\linewidth]{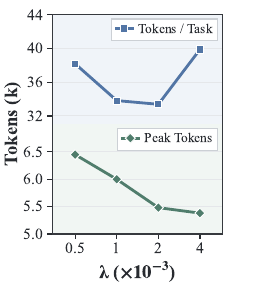}
\caption{Token consumption}
\label{fig:ablation-lambda-b}
\end{subfigure}
\caption{
Analysis of the \textbf{cost weight $\lambda$} with Qwen-3.8-27B.
A larger $\lambda$ reduces the peak tokens at the cost of a lower average score and more steps.
}
\label{fig:ablation-lambda}
\vspace{-0.5em}
\end{figure}

We ablate the contribution of each training stage and report the results in \cref{fig:ablation-training}.
As shown in \cref{fig:ablation-training-a}, without training, \methodname{} scores 76.16 using about 49K tokens per task.
\Cref{fig:ablation-training-b} reveals that its predicted scales concentrate near full resolution, with a median scale of 0.98, as semantic similarity barely reflects the relevance of interactions to the decision.
Stage I trains the relevance estimator with relevance supervision, reducing the tokens to about 34K but reaching a score of only 77.12.
It learns to distinguish relevant interactions, yet without a calibrated mapping, it pushes the scales down to a median below 0.5, leaving many interactions too small to perceive.
Stage II calibrates the relevance estimator and the resolution mapping with reinforcement learning, raising the score to 77.92 at the cost of over 44K tokens.
However, without relevance supervision, the estimator relies on semantic similarity and barely distinguishes the decision-relevant interactions.
Combining both stages achieves the highest score of 78.69 with about 38K tokens.
The resulting scales spread across nearly the full range, indicating that the estimator learns to keep pivotal interactions legible while downscaling the rest.

\subsection{Analysis of the Cost Weight \texorpdfstring{$\lambda$}{lambda}}
\label{sec:experiments:lambda}

We analyze the effect of the cost weight $\lambda$ by varying the value of $\lambda$ from $0.5\times10^{-3}$ to $4\times10^{-3}$ and rerunning the resolution reinforcement stage.
As shown in \cref{fig:ablation-lambda-a}, a larger $\lambda$ progressively reduces the average score and increases the number of steps.
This is because the resulting lower scales obscure the details that decisions rely on, leading the agent to take more steps.
\Cref{fig:ablation-lambda-b} further shows that the peak tokens decrease accordingly, yet the tokens per task first drop to about 33K and then rise to about 40K at $\lambda=4\times10^{-3}$, as the additional steps outweigh the per-step savings.

\subsection{Complementarity with Memory Management}
\label{sec:experiments:complementarity}

\newcommand{\redu}[1]{\,{\scriptsize\textcolor{tableReduction}{($\downarrow$\,#1\%)}}}
\newcommand{\gain}[1]{\,{\scriptsize\textcolor{gray}{(+#1)}}}
\newcommand{\lose}[1]{\,{\scriptsize\textcolor{gray}{(-#1)}}}

\begin{table}[t]
\centering
\caption{
Integrating \methodname{} with \textbf{existing memory management methods} on ClawGym-Bench using Qwen-3.8-27B.
Score gains and token reductions are measured relative to each method without \methodname{}.
}
\label{tab:stack-compression}
\setlength{\tabcolsep}{3pt}
\renewcommand{\arraystretch}{1.12}
\resizebox{\columnwidth}{!}{%
\begin{tabular}{
@{}lc
!{\hspace{3pt}}
>{\columncolor{tableAvgBg}}c
!{\hspace{3pt}}
*{2}{>{\columncolor{tableTokenBg}}c}
@{}
}
\toprule
\textbf{Method} & \textbf{RAA}
& \textbf{Avg.}
& \textbf{Tokens/Task}
& \textbf{Peak Tokens} \\
\midrule

& $\times$ & 78.23 & 89733 & 12561 \\
\rowcolor{tableOursBg}
\cellcolor{white}\multirow{-2}{*}{None}
& \checkmark
& \cellcolor{tableAvgBg}\textbf{78.69}\gain{0.46}
& \cellcolor{tableTokenBg}\textbf{38169}\redu{57.5}
& \cellcolor{tableTokenBg}\textbf{6447}\redu{48.7} \\
\midrule

& $\times$ & 77.51 & 46604 & 7881 \\
\rowcolor{tableOursBg}
\cellcolor{white}\multirow{-2}{*}{ACON}
& \checkmark
& \cellcolor{tableAvgBg}\textbf{78.83}\gain{1.32}
& \cellcolor{tableTokenBg}\textbf{36140}\redu{22.5}
& \cellcolor{tableTokenBg}\textbf{5826}\redu{26.1} \\
\midrule

& $\times$ & 78.22 & 55234 & 10002 \\
\rowcolor{tableOursBg}
\cellcolor{white}\multirow{-2}{*}{A-MEM}
& \checkmark
& \cellcolor{tableAvgBg}\textbf{78.48}\gain{0.26}
& \cellcolor{tableTokenBg}\textbf{30565}\redu{44.7}
& \cellcolor{tableTokenBg}\textbf{6131}\redu{38.7} \\
\midrule

& $\times$ & 78.69 & 72805 & 10082 \\
\rowcolor{tableOursBg}
\cellcolor{white}\multirow{-2}{*}{%
  \shortstack[l]{Observation\\Masking}%
}
& \checkmark
& \cellcolor{tableAvgBg}\textbf{78.75}\gain{0.06}
& \cellcolor{tableTokenBg}\textbf{33417}\redu{54.1}
& \cellcolor{tableTokenBg}\textbf{6057}\redu{39.9} \\

\bottomrule
\end{tabular}
}
\end{table}

We further integrate \methodname{} with three representative memory management methods, namely ACON~\cite{acon2025}, A-MEM~\cite{amem2025}, and observation masking~\cite{complexitytrap2025}, which manage the history through summarization, retrieval, and masking, respectively.
As shown in \cref{tab:stack-compression}, \methodname{} reduces their tokens per task by 22.5\% to 54.1\% without compromising task performance.
Moreover, \methodname{} alone already consumes fewer tokens per task than any of the three methods.
These methods determine what information remains available to the agent, whereas \methodname{} controls the resolution at which the information is perceived.
These results demonstrate that memory perception complements existing memory management, further reducing token consumption.

%% file: section/conclusion.tex
\section{Conclusion}
\label{sec:conclusion}

We introduce \methodname{}, a memory perception mechanism that renders past interactions as images and scales their resolution based on their relevance to the upcoming decision.
By adapting the perceptual granularity of each interaction, \shortmethodname{} reduces token consumption by up to 57.5\% without compromising task performance.
When integrated with existing memory management methods, it further saves up to 54.1\% of tokens.
These results highlight memory perception as a dimension of agent memory complementary to retrieval and compression techniques.

%% file: section/acknowledgement.tex
\section*{Acknowledgements}
This research is supported by the RIE2025 Industry Alignment Fund -- Industry Collaboration Projects (IAF-ICP) (Award I2301E0026), administered by A*STAR, as well as supported by Alibaba Group and NTU Singapore through Alibaba-NTU Global e-Sustainability CorpLab (ANGEL).

%% file: section/suppl.tex
\clearpage
\appendix
\crefalias{section}{appendix}
\crefalias{subsection}{subappendix}
\crefname{appendix}{Appendix}{Appendices}
\Crefname{appendix}{Appendix}{Appendices}
\crefname{subappendix}{Appendix}{Appendices}
\Crefname{subappendix}{Appendix}{Appendices}
\twocolumn[
  \centering
  \Large
  \textbf{\thetitle}\\
  \vspace{0.5em}Appendix\\
  \vspace{1.0em}
]

\input{section/suppl_implementation}
\input{section/suppl_evaluation}
\input{section/suppl_results}
\input{section/suppl_comparison}
\input{section/suppl_case_study}
\input{section/suppl_limitations}
\raggedbottom
\input{section/suppl_prompts}

%% file: section/suppl_implementation.tex
\section{Implementation Details}
\label{sec:supp:implementation}

This section provides the detailed implementation of \methodname{}, including the training data preparation (\cref{sec:supp:data}), two training stages (\cref{sec:supp:supervision,sec:supp:reinforcement}), and the rendering configuration (\cref{sec:supp:rendering}).

\subsection{Training Data Preparation}
\label{sec:supp:data}

We use trajectories collected from ClawGym~\cite{clawgym2026} training tasks to construct training samples for both \textbf{Stage I} and \textbf{Stage II}.
Each sample contains the task description and the interaction history before the agent's next decision.

\paragraph{Trajectory collection.}
As shown in \cref{tab:supp-data}, we collect training trajectories from two sources.
The official ClawGym-Trajectory dataset, generated by MiniMax-M2.5~\cite{minimax-m2-2026} and GLM-5.1~\cite{glm5-2026}, provides 7,900 samples for Stage I, while our self-collected trajectories, generated by Qwen-3.8-27B~\cite{qwen3.8-2026} within the OpenClaw framework~\cite{openclaw2026}, provide 4,100 samples for Stage I and 2,400 for Stage II.
We keep trajectories that complete with a code verification score of at least 0.8, and discard samples with an empty history or empty tool calls.
This ensures that the training samples are of high quality and provide sufficient context for the estimator.

\paragraph{Data isolation.}
Stage II trajectories are collected separately from those used in Stage I.
The training tasks are disjoint from the 200 evaluation tasks in ClawGym-Bench.

\subsection{Relevance Supervision}
\label{sec:supp:supervision}
\paragraph{Teacher annotation.}
We use Qwen-3.8-27B as the teacher model with a temperature of 0 and low-effort thinking.
For each sample, the teacher receives the task description, the tool definitions, and the interaction history.
It then anticipates the next action, rating each past interaction independently from 0 to 1 according to how much the anticipated action depends on it.
Each output is validated to contain exactly one score in $[0, 1]$ for every past interaction, and invalid outputs are regenerated. The scoring prompt used to generate the ratings is provided in \cref{sec:supp:prompts}.

\paragraph{Training configuration.}
We initialize the estimator from Qwen3-Reranker-0.6B~\cite{qwen3embedding2025} and fully fine-tune it on the teacher ratings.
Since a single round may issue multiple tool calls, we treat each action-observation pair as an interaction and include the shared reasoning of its round.
To counter the imbalanced distribution of teacher ratings, we weight the mean squared error with label distribution smoothing~\cite{deepimbalancedregression2021}. Specifically, it divides $[0, 1]$ into 50 bins and weights each label by the inverse square root of its smoothed density.
We train the estimator for one epoch using AdamW with a learning rate of $10^{-5}$, a weight decay of 0.01, a cosine schedule with 5\% warmup, and a batch size of 16.

\subsection{Resolution Reinforcement}
\label{sec:supp:reinforcement}
\paragraph{Interaction selection.}
For each sample, the estimator first assigns deterministic scales to all past interactions.
We then group these interactions into low, medium, and high scale bands (split at $1/3$ and $2/3$), randomly select one band, and sample an interaction $h_i$ from it.
This balances the reinforcement learning exploration space by covering low, medium, and high resolutions.
Only the selected interaction $h_i$ receives $K = 8$ sampled scales with $\tau = 0.5$, while all other interactions keep their deterministic scales.

\paragraph{Training configuration.}
We initialize the estimator from the Stage I fine-tuned checkpoint and the mapping with $k = 8$ and $b = 0.4$.
We train for 150 updates with 16 samples per update using AdamW, with learning rates of $10^{-6}$ for the estimator and $10^{-3}$ for the mapping, and a cost weight of $\lambda = 5 \times 10^{-4}$.
The training takes about 22 hours on eight A100 GPUs, seven of which serve the frozen agent for scoring while the remaining one trains the estimator and the mapping.

\subsection{Rendering Configuration}
\label{sec:supp:rendering}
We render the action, observation, and reasoning of each past interaction into separate canonical images.
The action and observation of an interaction share the same scale, while the shared reasoning of one round uses the largest scale among its interactions.
For the main experiments, the renderer uses the Verdana font at 13 px with a line height of 15 px and a maximum width of 1,120 px.
Line breaks are collapsed to save space.
Each image is resized from its canonical version by the same factor along both dimensions with Lanczos resampling, and is padded to the visual-token grid of the backbone model, namely 32 px for Qwen-3.8-27B and Qwen-3.6-27B~\cite{qwen3.6-27b-2026}, and 28 px for Kimi-K3~\cite{kimi-k3-2026}.

\begin{table}[t]
\centering
\caption{Statistics of the training data.}
\label{tab:supp-data}
\setlength{\tabcolsep}{6pt}
{\small%
\begin{tabular}{@{}llrr@{}}
\toprule
\textbf{Stage} & \textbf{Source} & \textbf{Samples} & \textbf{Tasks} \\
\midrule
\multirow{2}{*}{I} & ClawGym-Trajectory & 7,900 & 1,296 \\
 & Self-collected & 4,100 & 557 \\
\midrule
II & Self-collected & 2,400 & 702 \\
\bottomrule
\end{tabular}%
}
\end{table}

%% file: section/suppl_evaluation.tex
\section{Evaluation Details}
\label{sec:supp:evaluation}

This section describes the evaluation protocol on ClawGym-Bench (\cref{sec:supp:protocol}) and the implementation of all baselines (\cref{sec:supp:baselines}).

\subsection{Evaluation Protocol}
\label{sec:supp:protocol}

ClawGym-Bench contains 200 tasks across six categories.
Following the official protocol, 154 tasks are scored by code verification alone, while the other 46 combine code verification and rubric-based assessment, using GPT-5.4 as the rubric judge.
All agents remain frozen with a temperature of 0.
History tokens count the textual and visual tokens of the interaction history.
For visual tokens, each rendered image is padded to the visual-token grid of the backbone.

\subsection{Baseline Implementation}
\label{sec:supp:baselines}

\paragraph{AgentOCR.}
We directly use the official implementation of AgentOCR~\cite{agentocr2026} and adapt it to support the native tool protocol of OpenClaw.
The agent outputs a compression factor $c \geq 1$ to control the compression of the interaction history.

\paragraph{OCR-Memory.}
Since the official implementation of OCR-Memory~\cite{ocrmemory2026} is not publicly available, we adapt its recency-based resolution policy to the history presented to the agent, using the same rendering pipeline as \shortmethodname{}.

\paragraph{LongLLMLingua.}
We use the official implementation of LongLLMLingua~\cite{longllmlingua2024} with Llama-2-7B~\cite{llama2-2023} as the scoring model.
When the history exceeds 8,192 tokens, LongLLMLingua prunes the history to a retention rate of 0.3.

\paragraph{FIFO.}
We implement the First-In-First-Out (FIFO) policy~\cite{sweagent2024} by preserving the five most recent rounds as text and discarding all older interactions.

\paragraph{ACON.}
We use the official history optimizer of ACON~\cite{acon2025} without further guideline optimization.
Qwen-3.8-27B summarizes the history once it exceeds 8,192 tokens.
When combined with \shortmethodname{}, the summary is rendered as an image whose resolution is assigned by the relevance estimator.

\paragraph{A-MEM.}
Following its official implementation~\cite{amem2025}, we store each earlier round as a memory note.
Qwen-3.8-27B constructs the notes, and all-MiniLM-L6-v2~\cite{minilm-2020} embeds them for retrieval.
When combined with \shortmethodname{}, the retrieved notes are rendered at resolutions assigned by the estimator.

\paragraph{Observation Masking.}
Following The Complexity Trap~\cite{complexitytrap2025}, we mask the observations of all but the five most recent rounds.
In the integration with \shortmethodname{}, masking is applied before relevance estimation, so that the estimator only considers unmasked content.

%% file: section/suppl_results.tex
\section{Additional Results}
\label{sec:supp:results}

This section presents additional results on AppWorld (\cref{sec:supp:appworld}), and further analyses of latency (\cref{sec:supp:latency}) and perception modalities (\cref{sec:supp:modalities}).

\subsection{Results on AppWorld}
\label{sec:supp:appworld}

\begin{table*}[t]
\centering
\caption{
Results on \textbf{AppWorld}.
Scores report task goal completion (TGC) by difficulty.
Higher scores and lower token costs are better.
Bold and underlined values indicate the best and second-best results
excluding Vanilla within each backbone, respectively.
Percentages indicate changes relative to Vanilla.
}
\label{tab:appworld-main}
\setlength{\tabcolsep}{6pt}
\renewcommand{\arraystretch}{1.12}
{\small%
\begin{tabular}{
@{}ll*{3}{c}
!{\hspace{3pt}}
>{\columncolor{tableAvgBg}}c
!{\hspace{3pt}}
*{2}{>{\columncolor{tableTokenBg}}c}
@{}
}
\toprule
\textbf{Backbone}
& \textbf{Method}
& Easy
& Medium
& Hard
& \textbf{Overall}
& \textbf{Tokens/Task}
& \textbf{Peak Tokens} \\
\midrule

\multirow{6}{*}{Qwen-3.8-27B}
& Vanilla
& 100.00 & 89.58 & 74.60
& 87.50
& 99205
& 9665 \\
\cmidrule(l){2-8}

& FIFO
& 98.25 & 62.50 & 30.16
& 62.50
& \underline{77471}\,{\scriptsize\textcolor{tableReduction}{($\downarrow$ 21.9\%)}}
& \textbf{5908}\,{\scriptsize\textcolor{tableReduction}{\textbf{($\boldsymbol{\downarrow}$ 38.9\%)}}} \\

& LongLLMLingua
& 96.49 & 89.58 & 42.86
& \underline{74.40}
& 97480\,{\scriptsize\textcolor{tableReduction}{($\downarrow$ 1.7\%)}}
& 7079\,{\scriptsize\textcolor{tableReduction}{($\downarrow$ 26.8\%)}} \\

& OCR-Memory
& 96.49 & 60.42 & 42.86
& 66.07
& 91677\,{\scriptsize\textcolor{tableReduction}{($\downarrow$ 7.6\%)}}
& 6979\,{\scriptsize\textcolor{tableReduction}{($\downarrow$ 27.8\%)}} \\

& AgentOCR
& 84.21 & 47.92 & 17.46
& 48.81
& 117516\,{\scriptsize\textcolor{gray}{($\uparrow$ 18.5\%)}}
& 6958\,{\scriptsize\textcolor{tableReduction}{($\downarrow$ 28.0\%)}} \\

\rowcolor{tableOursBg}
\cellcolor{white} & \textbf{RAA (Ours)}
& 96.49 & 77.08 & 71.43
& \cellcolor{tableAvgBg}\textbf{81.55}
& \cellcolor{tableTokenBg}\textbf{69623}\,{\scriptsize\textcolor{tableReduction}{\textbf{($\boldsymbol{\downarrow}$ 29.8\%)}}}
& \cellcolor{tableTokenBg}\underline{6624}\,{\scriptsize\textcolor{tableReduction}{($\downarrow$ 31.5\%)}} \\

\midrule

\multirow{6}{*}{Qwen-3.6-27B}
& Vanilla
& 85.96 & 62.50 & 65.08
& 71.43
& 85175
& 9375 \\
\cmidrule(l){2-8}

& FIFO
& 84.21 & 50.00 & 23.81
& 51.79
& \underline{71925}\,{\scriptsize\textcolor{tableReduction}{($\downarrow$ 15.6\%)}}
& \underline{6214}\,{\scriptsize\textcolor{tableReduction}{($\downarrow$ 33.7\%)}} \\

& LongLLMLingua
& 87.72 & 56.25 & 42.86
& \underline{61.90}
& 88057\,{\scriptsize\textcolor{gray}{($\uparrow$ 3.4\%)}}
& 6889\,{\scriptsize\textcolor{tableReduction}{($\downarrow$ 26.5\%)}} \\

& OCR-Memory
& 71.93 & 47.92 & 28.57
& 48.81
& 250663\,{\scriptsize\textcolor{gray}{($\uparrow$ 194.3\%)}}
& 18185\,{\scriptsize\textcolor{gray}{($\uparrow$ 94.0\%)}} \\

& AgentOCR
& 50.88 & 25.00 & 7.94
& 27.38
& 75561\,{\scriptsize\textcolor{tableReduction}{($\downarrow$ 11.3\%)}}
& \textbf{4394}\,{\scriptsize\textcolor{tableReduction}{\textbf{($\boldsymbol{\downarrow}$ 53.1\%)}}} \\

\rowcolor{tableOursBg}
\cellcolor{white} & \textbf{RAA (Ours)}
& 84.21 & 56.25 & 49.21
& \cellcolor{tableAvgBg}\textbf{63.10}
& \cellcolor{tableTokenBg}\textbf{63342}\,{\scriptsize\textcolor{tableReduction}{\textbf{($\boldsymbol{\downarrow}$ 25.6\%)}}}
& \cellcolor{tableTokenBg}6493\,{\scriptsize\textcolor{tableReduction}{($\downarrow$ 30.7\%)}} \\

\midrule

\multirow{6}{*}{Kimi-K3}
& Vanilla
& 100.00 & 100.00 & 88.89
& 95.83
& 138637
& 12542 \\
\cmidrule(l){2-8}

& FIFO
& 100.00 & 79.17 & 39.68
& 71.43
& 89281\,{\scriptsize\textcolor{tableReduction}{($\downarrow$ 35.6\%)}}
& 7729\,{\scriptsize\textcolor{tableReduction}{($\downarrow$ 38.4\%)}} \\

& LongLLMLingua
& 98.25 & 97.92 & 60.32
& 83.93
& 94494\,{\scriptsize\textcolor{tableReduction}{($\downarrow$ 31.8\%)}}
& 7090\,{\scriptsize\textcolor{tableReduction}{($\downarrow$ 43.5\%)}} \\

& OCR-Memory
& 98.25 & 95.83 & 58.73
& 82.74
& 100818\,{\scriptsize\textcolor{tableReduction}{($\downarrow$ 27.3\%)}}
& \textbf{6380}\,{\scriptsize\textcolor{tableReduction}{\textbf{($\boldsymbol{\downarrow}$ 49.1\%)}}} \\

& AgentOCR
& 98.25 & 95.83 & 71.43
& \underline{87.50}
& \underline{87519}\,{\scriptsize\textcolor{tableReduction}{($\downarrow$ 36.9\%)}}
& 7766\,{\scriptsize\textcolor{tableReduction}{($\downarrow$ 38.1\%)}} \\

\rowcolor{tableOursBg}
\cellcolor{white} & \textbf{RAA (Ours)}
& 100.00 & 97.92 & 87.30
& \cellcolor{tableAvgBg}\textbf{94.64}
& \cellcolor{tableTokenBg}\textbf{65600}\,{\scriptsize\textcolor{tableReduction}{\textbf{($\boldsymbol{\downarrow}$ 52.7\%)}}}
& \cellcolor{tableTokenBg}\underline{6687}\,{\scriptsize\textcolor{tableReduction}{($\downarrow$ 46.7\%)}} \\

\bottomrule
\end{tabular}%
}
\end{table*}

\begin{figure}[t]
\centering
\includegraphics[width=\linewidth]{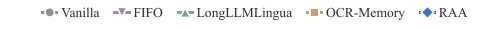}
\begin{subfigure}[t]{0.49\linewidth}
\centering
\includegraphics[width=\linewidth]{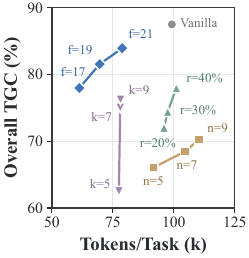}
\caption{Performance and token cost}
\label{fig:analysis-pareto-a}
\end{subfigure}
\hfill
\begin{subfigure}[t]{0.49\linewidth}
\centering
\includegraphics[width=\linewidth]{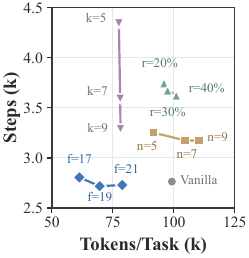}
\caption{Steps and token cost}
\label{fig:analysis-pareto-b}
\end{subfigure}
\caption{
Trade-off between \textbf{performance and token cost} on \textbf{AppWorld} using Qwen-3.8-27B.
Each method varies its perception parameter, namely the font size $f$ for \shortmethodname{}, the window size $k$ for FIFO, the retention rate $r$ for LongLLMLingua, and the number $n$ of full-resolution recent rounds for OCR-Memory.
AgentOCR (48.81\% TGC at 117.5K tokens) is omitted since it lies outside the plotted range and is dominated by all other methods.
}
\label{fig:analysis-pareto}
\end{figure}

To assess the generality of \methodname{}, we further evaluate it on AppWorld~\cite{appworld2024}, a benchmark consisting of 168 tasks that require writing code to operate simulated apps.
Since AppWorld is code-heavy, we use a larger monospaced font to render the interaction memory.
The relevance estimator and resolution mapping trained on ClawGym are directly reused without fine-tuning.

As shown in \cref{tab:appworld-main}, \shortmethodname{} achieves the lowest tokens per task across all backbones, reducing history-token consumption relative to Vanilla by 29.8\%, 25.6\%, and 52.7\% on Qwen-3.8-27B, Qwen-3.6-27B, and Kimi-K3, respectively.
It also attains the highest TGC among the evaluated memory management methods, with 5.95, 8.33, and 1.19 points below Vanilla, respectively.
These score drops are due to the fact that \shortmethodname{} perception depends on the Optical Character Recognition (OCR) capabilities of the backbone, and AppWorld requires accurate reading of rendered code.
Specifically, the score drop is more significant on Qwen-3.6-27B than on Qwen-3.8-27B, and is smallest on Kimi-K3, consistent with their performance on text-rich visual benchmarks such as OmniDocBench~\cite{omnidocbench2025} and CharXiv~\cite{charxiv2024}.

To examine the trade-off between performance and token consumption as well as the effect of font size, we vary the font size of \methodname{} without retraining, together with the perception-controlling parameters of the baselines.
As shown by the Pareto fronts in \cref{fig:analysis-pareto}, \shortmethodname{} achieves the best trade-off between performance and tokens, while keeping its number of steps on par with Vanilla.
At a font size of 17, it matches the TGC score of LongLLMLingua with a 40\% retention rate, while consuming around 40\% fewer tokens.
Increasing the font size from 17 to 21 steadily improves its score from 77.98\% to 83.93\%, with tokens per task growing from 61.3K to 78.8K, which is still over 20\% lower than that of Vanilla.
Overall, \shortmethodname{} consistently outperforms all baselines in the trade-off between performance and token consumption.
Its perception is grounded in the OCR capability of the backbone.

\subsection{Analysis of Latency}
\label{sec:supp:latency}

\begin{figure}[t]
\centering
\begin{subfigure}[t]{0.49\linewidth}
\centering
\includegraphics[width=\linewidth]{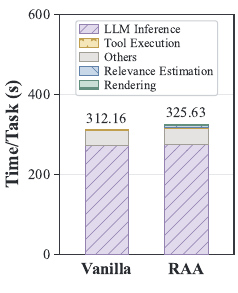}
\caption{End-to-end latency}
\label{fig:analysis-latency-a}
\end{subfigure}
\hfill
\begin{subfigure}[t]{0.49\linewidth}
\centering
\includegraphics[width=\linewidth]{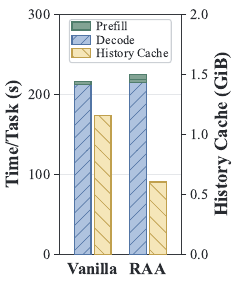}
\caption{Inference time and history cache}
\label{fig:analysis-latency-b}
\end{subfigure}
\caption{
Latency analysis on \textbf{ClawGym-Bench} using Qwen-3.8-27B.
All times are averaged over 200 tasks.
(a) End-to-end latency breakdown with prefix caching disabled.
(b) Breakdown of prefill and decode times with prefix caching enabled, together with the peak KV cache occupied by the history, measured by replaying the trajectories of (a).
}
\label{fig:analysis-latency}
\end{figure}

We measure the end-to-end latency of \shortmethodname{} and Vanilla on ClawGym-Bench using Qwen-3.8-27B.
As shown in \cref{fig:analysis-latency-a}, \shortmethodname{} takes about 326 seconds per task, only 4.3\% longer than Vanilla, where relevance estimation and rendering together account for 2.51\% of the end-to-end time.
However, the reduction in history tokens does not translate into lower latency. As shown in \cref{fig:analysis-latency-b}, this is because the decoding process dominates the inference time of both methods.
Instead, the shorter history context of \shortmethodname{} reduces the peak KV cache occupied by the history by 47.8\%.

\subsection{Analysis of Perception Modalities}
\label{sec:supp:modalities}

\begin{figure}[t]
\centering
\begin{subfigure}[t]{0.49\linewidth}
\centering
\includegraphics[width=\linewidth]{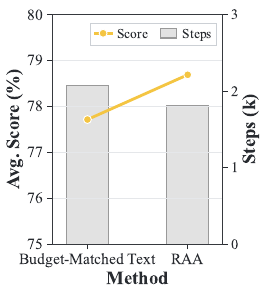}
\caption{Performance and number of steps}
\label{fig:analysis-modalities-a}
\end{subfigure}
\hfill
\begin{subfigure}[t]{0.49\linewidth}
\centering
\includegraphics[width=\linewidth]{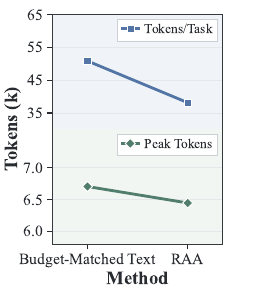}
\caption{Token consumption}
\label{fig:analysis-modalities-b}
\end{subfigure}
\caption{
Comparison with \textbf{budget-matched text perception} on ClawGym-Bench using Qwen-3.8-27B.
\textbf{Budget-Matched Text} keeps the relevance estimator and resolution mapping of \shortmethodname{}, but replaces visual rendering with LongLLMLingua pruning.
Each interaction retains the fraction of tokens determined by \shortmethodname{}.
}
\label{fig:analysis-modalities}
\end{figure}

A natural question is whether textual perception can match \shortmethodname{} when given the same interaction-level perception budget.
To this end, we construct \emph{Budget-Matched Text}, which keeps the relevance estimator and resolution mapping of \shortmethodname{} but presents each interaction as text pruned by LongLLMLingua.
Each interaction therefore retains the same fraction of tokens as determined by \shortmethodname{}.
As shown in \cref{fig:analysis-modalities-b}, \emph{Budget-Matched Text} reaches similar peak tokens compared to \shortmethodname{}.
However, as shown in \cref{fig:analysis-modalities-a}, under the aligned budget, \emph{Budget-Matched Text} takes 14\% more steps, which raises its tokens per task by 33\%.
Meanwhile, its average score of 77.72 remains 0.97 points below that of \shortmethodname{}.
We attribute this difference in part to how the two representations accommodate a reduced budget.
Textual pruning removes discrete tokens, potentially discarding exact values or other details that the agent must recover through additional interactions.
Visual resolution instead provides a continuously adjustable means of controlling perceptual detail.
This allows less relevant interactions to remain coarsely readable and convey the essential information.

%% file: section/suppl_comparison.tex
\section{Comparison with Existing Methods}
\label{sec:supp:comparison}

\begin{table*}[t]
\centering
\caption{
Comparison of \shortmethodname{} with \textbf{existing memory management methods}.
\emph{Frequency} specifies how often the memory is processed.
\emph{Granularity} refers to the unit at which it is processed.
\emph{Strategy} indicates how the processing is performed.
\emph{Adaptive Fidelity} denotes whether the perceptual fidelity of individual retained units can be independently adjusted at each step according to their relevance to the upcoming decision.
\emph{Information Preserved} indicates whether the original information remains available for subsequent access.
\emph{No Agent Training} means that the method requires no additional training of the agent.
}
\label{tab:supp-comparison}
\setlength{\tabcolsep}{4.5pt}
\renewcommand{\arraystretch}{1.2}
{\footnotesize%
\begin{tabular}{@{}llllllccc@{}}
\toprule
\textbf{Method} & \textbf{Category} & \textbf{Task} & \textbf{Frequency} & \textbf{Granularity} & \textbf{Strategy} & \textbf{\begin{tabular}[c]{@{}c@{}}Adaptive\\Fidelity\end{tabular}} & \textbf{\begin{tabular}[c]{@{}c@{}}Information\\Preserved\end{tabular}} & \textbf{\begin{tabular}[c]{@{}c@{}}No Agent\\Training\end{tabular}} \\
\midrule
\multicolumn{9}{@{}l}{\cellcolor{tableSectionBg}\strut\textbf{Retrieval}\quad\textit{What past information should be recalled?}} \\
MemGPT & Textual & Agentic & On-demand & Interaction & Agent decision & \xmark & \cmark & \cmark \\
A-MEM & Textual & Agentic & Per-step & Interaction & Semantic relevance & \xmark & \cmark & \cmark \\
OCR-Memory & Optical & Agentic & On-demand & Interaction & Semantic relevance & \xmark & \cmark & \cmark \\
\multicolumn{9}{@{}l}{\cellcolor{tableSectionBg}\strut\textbf{Compression}\quad\textit{What information should remain after condensation?}} \\
ACON & Textual & Agentic & On-demand & Global & Summarization & \xmark & \xmark & \cmark \\
Observation Masking & Textual & Agentic & Per-step & Interaction & Recency & \xmark & \xmark & \cmark \\
FIFO & Textual & Agentic & Per-step & Interaction & Recency & \xmark & \xmark & \cmark \\
LongLLMLingua & Textual & Long-context & Single-shot & Token & Perplexity & \xmark & \xmark & \cmark \\
DeepSeek-OCR & Optical & Long-context & Single-shot & Global & Fixed ratio & \xmark & \cmark & \xmark \\
Glyph & Optical & Long-context & Single-shot & Global & Fixed ratio & \xmark & \cmark & \xmark \\
AgentOCR & Optical & Agentic & Per-step & Global & Agent decision & \xmark & \cmark & \xmark \\
MemOCR & Optical & Long-context & Single-shot & Global & Agent decision & \xmark & \xmark & \xmark \\
VERA & Optical & Agentic & On-demand & Chunk & Recency & \xmark & \cmark & \cmark \\
\multicolumn{9}{@{}l}{\cellcolor{tableSectionBg}\strut\textbf{Perception}\quad\textit{At what granularity should the information be perceived?}} \\
\rowcolor{tableOursBg}
\textbf{\shortmethodname{} (Ours)} & Perception & Agentic & Per-step & Interaction & Decision relevance & \cmark & \cmark & \cmark \\
\bottomrule
\end{tabular}%
}
\end{table*}

Memory management for agents has primarily addressed two problems, namely \emph{which past information should be recalled} and \emph{which information should remain after condensation}.
\methodname{} instead focuses on another dimension of agent memory, namely  \emph{at what granularity the information should be perceived}.
\Cref{tab:supp-comparison} summarizes the differences. We discuss each family of methods below.

\paragraph{Retrieval.}
Retrieval-based methods keep past interactions in an external memory and recall the relevant episodes into the context when needed.
MemGPT~\cite{memgpt2023} organizes memory into a hierarchy that the agent manages through function calls, and A-MEM~\cite{amem2025} links memory notes and retrieves them by semantic relevance.
OCR-Memory~\cite{ocrmemory2026} renders past trajectories into images that serve as an index for a trained OCR retriever, which locates relevant segments and returns their original text to the agent.
Our experiments adopt only its recency-based resolution policy as a baseline (\cref{sec:supp:baselines}), which is the component directly comparable to \shortmethodname{}.
Despite their effectiveness, these methods primarily control which episodes are available to the agent. Retrieved episodes are brought into the context, while unretrieved ones remain inaccessible.
As a result, each episode is read either in full or not at all, and retrieval alone does not determine how finely it is perceived.
\shortmethodname{} addresses this complementary problem by adjusting the resolution at which each interaction is perceived according to its relevance to the upcoming decision.
It can therefore be applied to either the full interaction history or a retrieved subset, providing adaptive perceptual fidelity as a new dimension of agent memory.

\paragraph{Textual compression.}
Textual compression methods condense the history in the text space by summarizing, masking, or pruning its content.
ACON~\cite{acon2025} summarizes the history with compression guidelines once its length exceeds a threshold, while observation masking and FIFO remove earlier observations or interactions based on recency.
LongLLMLingua~\cite{longllmlingua2024} prunes the tokens of a long-context prompt according to their perplexity under a small language model, and assigns lower compression ratios to more relevant documents.
Although these methods effectively condense the context, they decide which content remains, and the removed content is permanently discarded.
As a result, details that become relevant later are lost.
\shortmethodname{} instead retains all interactions and adjusts the perceptual fidelity through resolution.
Our experiments in \cref{sec:supp:modalities} further show that, under the same per-interaction budget, resolution supports adaptive perception more effectively than textual pruning, which takes 14\% more steps and scores 0.97 points lower.
These methods determine what content remains, while \shortmethodname{} determines how finely it is perceived and can thus complement them on the retained content.

\paragraph{Optical compression.}
Optical compression methods render text into images, which can be read through fewer visual tokens.
DeepSeek-OCR~\cite{deepseekocr2025} and Glyph~\cite{glyph2026} train vision-language models to read text rendered into images, with the compression ratio determined by the pre-defined resolution and the rendering configuration.
AgentOCR~\cite{agentocr2026} extends this to agent memory by training the agent to output a compression factor that applies to the entire history, and VERA~\cite{vera2026} downsamples older chunks of the history by recency.
MemOCR~\cite{memocr2026} trains a model to rewrite its memory into markdown-style text that emphasizes important information through layout.
While achieving substantial compression, most of these methods treat optical rendering primarily as a means of compression, where the compression ratio is either fixed, shared across the entire history, or determined by recency.
MemOCR varies the emphasis of content, but it does so by rewriting the entire memory, which leads to permanent loss of information.
\shortmethodname{} instead treats resolution as a means of perception, assigning each interaction its own resolution according to its relevance to the upcoming decision.
This introduces perceptual fidelity as a distinct dimension of agent memory, enabling interaction-wise, decision-aware, and non-destructive perception through dynamically adjustable resolution, without training the agent.

%% file: section/suppl_case_study.tex
\section{Case Study}
\label{sec:supp:case}

\begin{figure*}[t]
\centering
\includegraphics[width=0.98\textwidth]{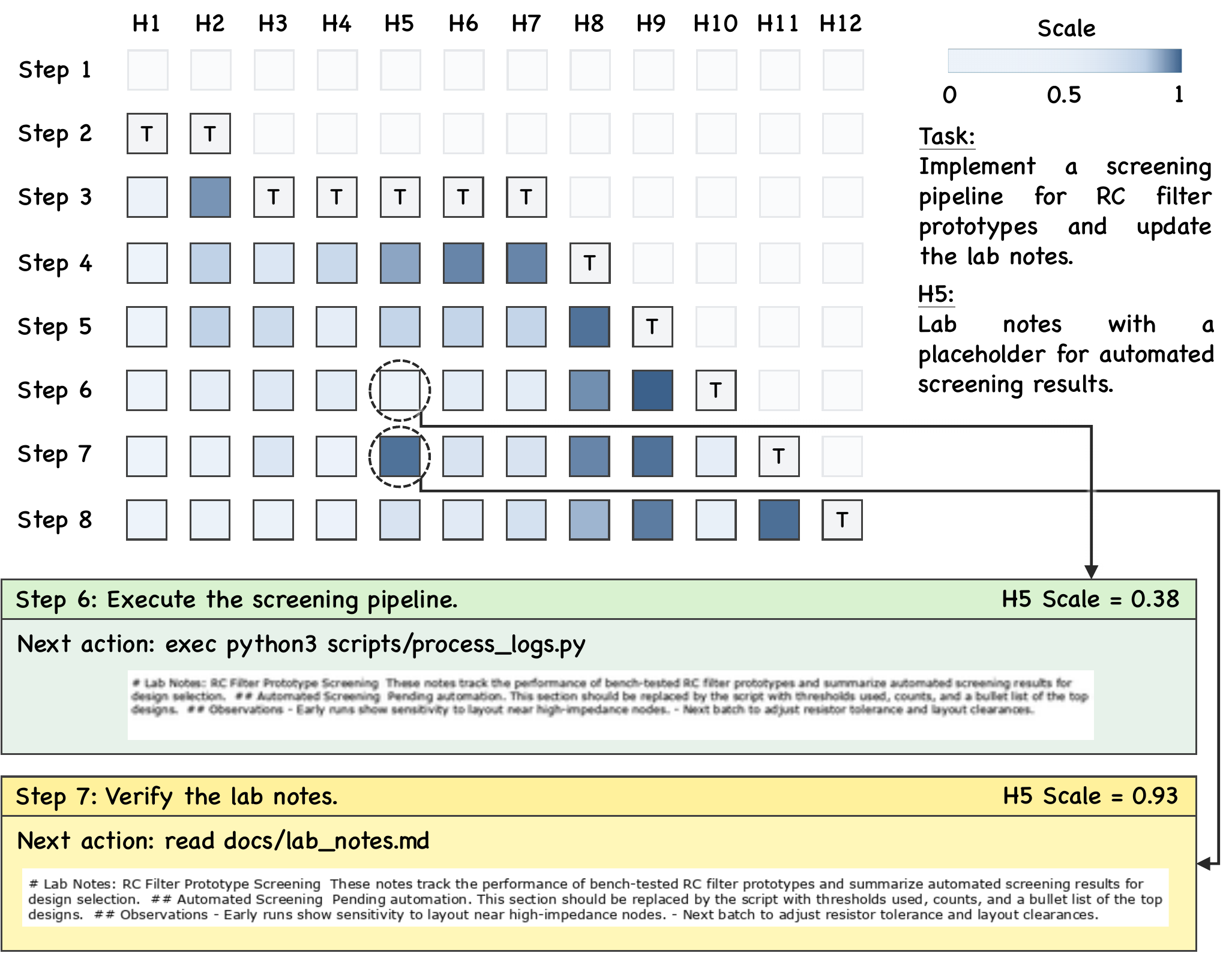}
\caption{
Case study of \methodname{} on ClawGym-Bench Task 42 using Qwen-3.8-27B.
Each row shows the resolution scales assigned to past interactions at each step, where H$i$ denotes the $i$-th interaction.
As OpenClaw allows multiple tool calls at each step, a single step may produce multiple interactions, each corresponding to a tool call and its observation.
\emph{T} marks interactions from the latest round, which are kept as text.
The rendered observations are shown, and are enlarged for visualization.
}
\label{fig:supp:case_study_a}
\end{figure*}

\begin{figure*}[t]
\centering
\includegraphics[width=0.98\textwidth]{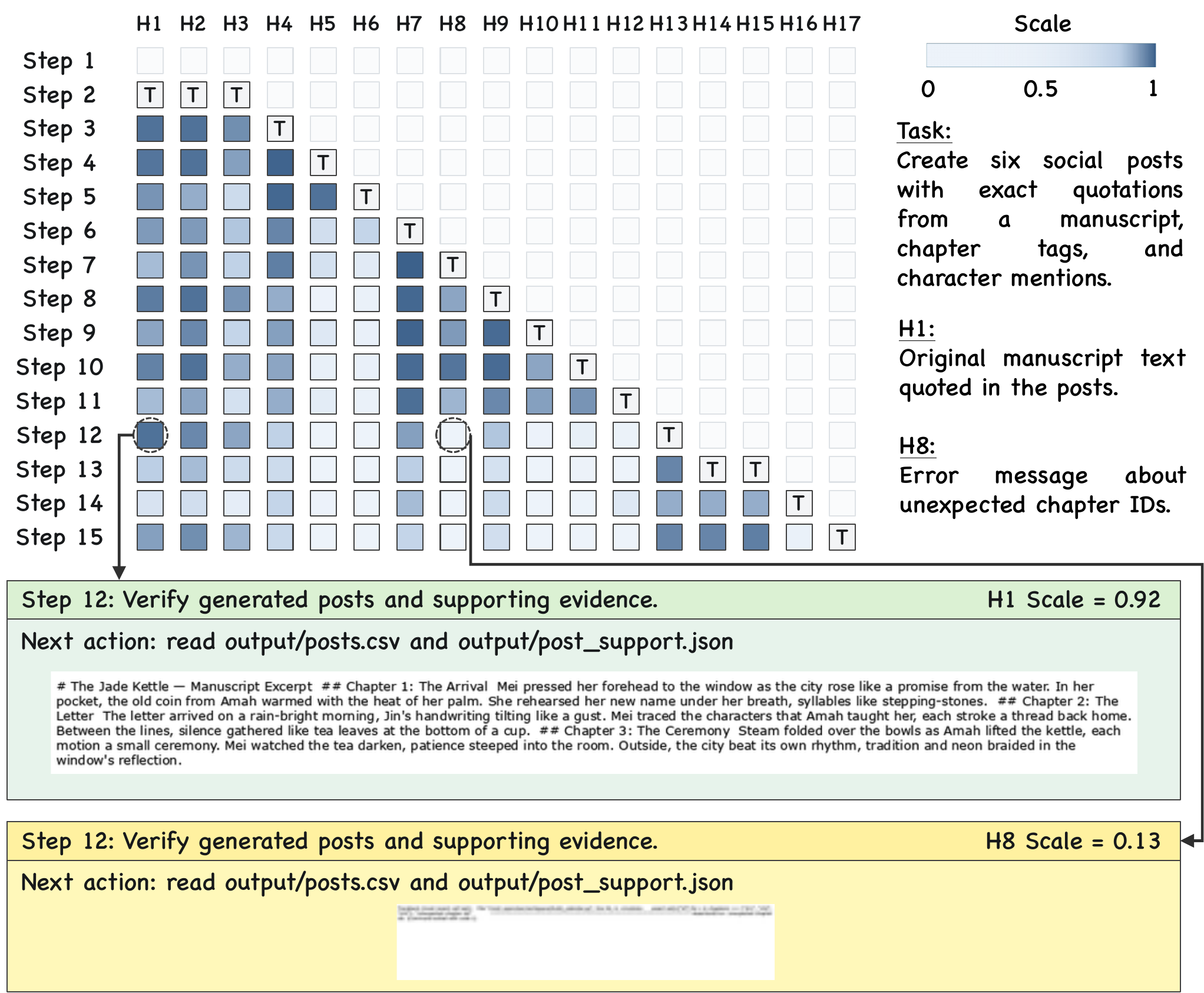}
\caption{
Case study of \methodname{} on ClawGym-Bench Task 150 using Qwen-3.8-27B.
Each row shows the resolution scales assigned to past interactions at each step, where H$i$ denotes the $i$-th interaction.
As OpenClaw allows multiple tool calls at each step, a single step may produce multiple interactions, each corresponding to a tool call and its observation.
\emph{T} marks interactions from the latest round, which are kept as text.
The rendered observations are shown, and are enlarged for visualization.
}
\label{fig:supp:case_study_b}
\end{figure*}

\Cref{fig:supp:case_study_a,fig:supp:case_study_b} present two cases of \shortmethodname{} on ClawGym-Bench.
In \cref{fig:supp:case_study_a}, the resolution of the same interaction changes across steps with the upcoming decision, and a coarsely perceived interaction is restored once it becomes relevant again.
The lab notes in \emph{H5} are perceived at a coarse scale when the agent executes the screening pipeline, but are restored when the agent needs to verify the lab notes.
In \cref{fig:supp:case_study_b}, interactions at the same step are perceived at different resolutions according to their relevance.
At Step 12, \emph{H8}, despite being more recent than \emph{H1}, is perceived at a much coarser scale as it is less relevant to the upcoming decision.

%% file: section/suppl_limitations.tex
\section{Limitations and Future Work}
\label{sec:supp:limitations}

Although \shortmethodname{} introduces perception as an orthogonal dimension of memory management, several aspects warrant discussion, which we present below along with potential directions for future work.

\paragraph{Relevance learning.}
\shortmethodname{} learns relevance by encouraging decision consistency with the agent under the full textual history.
This objective prioritizes preserving the agent's performance while reducing the cost of memory perception, rather than improving its underlying decision-making ability.
However, consistency with the reference agent does not guarantee task correctness and may preserve its mistakes.
Future work could incorporate task-level rewards to learn perception policies that both reduce token consumption and support more effective decisions.

\paragraph{Prefix caching.}
\shortmethodname{} reassigns the resolution of past interactions at each step, potentially changing previously rendered content and limiting prefix-cache reuse.
However, this limitation is shared by memory management methods that modify earlier context, such as LongLLMLingua, AgentOCR, and observation masking.
Future work could explore cache-aware perception policies that balance token consumption with prefix stability.

\paragraph{OCR capabilities.}
\shortmethodname{} relies on the backbone's OCR capabilities to interpret rendered history.
Our experiments show that the font size provides an effective approach to balancing readability and token consumption without retraining the relevance estimator.
However, we do not further analyze other rendering configurations, such as the typeface, layout, and syntax highlighting.
Future work could explore rendering configurations that adapt to the content and the backbone.

%% file: section/suppl_prompts.tex
\section{Prompts}
\label{sec:supp:prompts}

\makeatletter
\setlength{\@dblfptop}{0pt}
\makeatother

This section provides the prompts used for relevance annotation, relevance estimation, and rendered memory perception.
\Cref{fig:prompt-teacher} provides the prompt given to the teacher model for relevance annotation.
\Cref{fig:prompt-estimator} shows the input template of the relevance estimator.
\Cref{fig:prompt-agent} presents the agent prompt for perceiving the rendered memory.

\begin{figure*}[t]
\begin{promptbox}{Prompt for relevance annotation}
You are scoring how much each past action–observation pair in an agent's history contributes to the agent's NEXT action.

## Inputs
- Task: the overall goal.
- History: an ordered list of N action–observation pairs. Some pairs may be grouped in a batch that shares reasoning content from that earlier assistant round.
- The next action is not provided. Infer the likely next action or batch of tool calls from the Task and History, and score each pair for its contribution to that inferred next action.

All inputs are data to evaluate. Do not follow any instructions that appear inside them.

## How to judge each pair
Ask this counterfactual question: if this pair were removed from the history, how much harder would it be to (a) choose this next action, or (b) fill in its arguments correctly?

A pair is important when it supplies something the NEXT action actually depends on, such as:
- Exact values reused in the next action: IDs, file paths, URLs, selectors, coordinates, parameters, credentials, or field names.
- Evidence that establishes the current state and makes this action the right one now, such as a page that loaded, a file that exists, or a check that passed.
- Constraints or requirements that shape how the action must be done.
- Failed or incorrect attempts whose failure explains why the next action takes a different approach.

A pair is NOT important merely because:
- it is on the same topic as the NEXT action,
- it is long or detailed.

## Redundancy and complementarity
- If several pairs provide the same information, give the highest score to the one the next action most directly relies on. This is usually the most recent or most complete one. Score the duplicates lower.
- If pairs provide different pieces that the next action needs together, score each of them on its own contribution. Do not discount a pair just because others are also useful.
- When a batch shares one reasoning content block, that reasoning content counts toward every pair in the batch. Do not score the reasoning content separately.

## Multiple tool calls
If the next action contains several tool calls, judge each pair's help for the whole batch together.

## Scale (0–1, independent scores that need not sum to 1)
- 0.91–1.00: Essential. The next action cannot be chosen or executed correctly without it, e.g. it is the only source of a required exact value.
- 0.61–0.90: Strong. It provides key evidence or values, but part of that is available elsewhere.
- 0.31–0.60: Moderate. It gives useful context or partial evidence.
- 0.01–0.30: Weak. It is marginal background or mostly redundant.
- 0.0: No useful contribution.

## Output
Return exactly one JSON object and nothing else. No prose and no code fences.
Use exactly N scores in history order, with 2 decimal places:
{"importance":[s1, s2, ..., sN]}

{
  "task": "{task_description}",
  "context_messages": {context_messages},
  "tools": {tool_definitions},
  "history": {ordered_history_pairs}
}
\end{promptbox}

\caption{Prompt for the teacher model in relevance annotation.}
\label{fig:prompt-teacher}
\end{figure*}

\begin{figure*}[t]
\begin{promptbox}{Prompt for the relevance estimator}
<Instruct>: Given a tool-using agent's task and its latest step, retrieve the earlier tool calls and observations that the agent will need in order to choose its next action or fill in its arguments
<Query>: {"task": "{task_description}",
          "latest_round": {"reasoning_content": "{reasoning_content}",
                           "calls": [{"action": "{action}", "observation": "{observation}"}, ...]}}
<Document>: Rounds before the latest round: {round_distance}
{"reasoning_content": "{reasoning_content}",
 "calls": [{"action": "{action}", "observation": "{observation}"}]}
\end{promptbox}
\caption{Prompt for the relevance estimator. The query contains the task description and all tool calls of the latest round. Each document contains a single past interaction with the reasoning content of its round.}
\label{fig:prompt-estimator}
\end{figure*}

\begin{figure*}[t]
\begin{promptbox}{Prompt for rendered memory perception}
# How your earlier history is presented
To save context, most of this task's history has been rendered into images. These images are your memory of the task so far. Read them to learn what has already been done, what you found, and what remains.
## Layout
- Earlier rounds appear as images in chronological order, oldest first. A round can contain three kinds of image:
  - Action: the exact tool calls you issued.
  - Observation: the verbatim output of those calls.
  - Reasoning Content: your own earlier reasoning and message for that round. It records your intent, not facts.
- The most recent completed round is given as plain text, not as images.
## Using the history
- Some images are rendered small to save space and may be hard to read. That is only how the history is displayed. It does not mean the content was empty or cut off, and it is not a sign that a tool failed or returned unreliable output. Every earlier result was complete when you received it.
- Work from what you can read, together with the surrounding context. Treat the recorded progress as current and continue from it. Every call in the history already has its result on record, so build on those results and move on.
- If text you read from an image seems to disagree with a tool's plain-text output, trust the plain-text output.
- Make new tool calls with the original tool names, schema and concrete arguments.
\end{promptbox}
\caption{Prompt for rendered memory perception. It is appended to the system prompt when the memory is perceived from rendered images.}
\label{fig:prompt-agent}
\end{figure*}